\documentclass{article}

\usepackage[letterpaper,top=2cm,bottom=2cm,left=3cm,right=3cm,marginparwidth=2cm]{geometry}
\usepackage[USenglish]{babel}
\usepackage[colorlinks=true, allcolors=blue]{hyperref}
\usepackage{enumitem}
\usepackage{titlesec}

\usepackage{amsmath}
\usepackage{algpseudocode}
\usepackage{csquotes}

\usepackage{listings}
\usepackage{xcolor}

\lstdefinelanguage{ocaml}{
  keywords={type, let, fun, lemma, match, with, if, then, else},
  sensitive=true,
  comment=[n]{(*}{*)},
  morestring=[b]",
}

\usepackage{graphicx}
\usepackage{pgfplots}
\usepgfplotslibrary{statistics, polar}
\pgfplotsset{compat=newest,
boxplot-style/.style={
            boxplot,
            draw=blue,
            fill=blue!30!white,
            boxplot/every median/.style={red, very thick},
            mark=*,
            every mark/.append style={
                fill=orange,
                draw=black,
                opacity=0.5
            },
        },}
\usepackage{adjustbox}
\usepackage[justification=centering]{caption}
\usepackage{subcaption}
\usepackage{tikz}
\usetikzlibrary{arrows.meta,positioning,calc,fit,backgrounds}
\usepackage{xcolor}
\usepackage{amssymb}
\usepackage{pifont}
\tikzset{
    box/.style={rectangle, draw, rounded corners, minimum width=3.5cm, minimum height=1cm, align=center}
}

\usepackage[square, numbers]{natbib}
\addto{\captionsenglish}{}

\usepackage{tcolorbox}
\usepackage{listings}
\usepackage{booktabs}
\usepackage{amssymb}

\usepackage[colorinlistoftodos]{todonotes}
\usepackage{authblk}
\title{Imandra CodeLogician: Neuro-Symbolic Reasoning for Precise Analysis of Software Logic}

\author[1]{Hongyu Lin}
\author[1]{Samer Abdallah}
\author[1]{Makar Valentinov}
\author[1]{Paul Brennan}
\author[1]{Elijah Kagan}
\author[1]{Christoph M. Wintersteiger}
\author[1]{Denis Ignatovich}
\author[1,2]{Grant Passmore\thanks{The author would like to thank the Isaac Newton Institute for Mathematical Sciences, Cambridge, for support and hospitality during the programme \emph{Big Proof: Formalizing Mathematics at Scale}, where some of the work on this paper was undertaken. This work was supported by EPSRC grant EP/Z000580/1.}}

\affil[1]{Imandra Inc.}
\affil[2]{Clare Hall, University of Cambridge}

\affil[ ]{\texttt{\{hongyu,samer,makar,paul,elijah,christoph,denis,grant\}@imandra.ai}}

\begin{document}
\maketitle

\begin{abstract}
Large Language Models (LLMs) have shown strong performance on code understanding and software engineering tasks, yet they fundamentally lack the ability to perform precise, exhaustive mathematical reasoning about program behavior. 
Existing benchmarks either focus on mathematical proof automation, largely disconnected from real-world software, or on engineering tasks that do not require semantic rigor.

We present \emph{CodeLogician}\footnote{\url{https://www.codelogician.dev}}, a neurosymbolic agent and framework for precise analysis of software logic, and its integration with \emph{ImandraX}\footnote{\url{https://www.imandra.ai/core}}, an industrial automated reasoning engine successfully deployed in financial markets, safety-critical systems and government applications.
Unlike prior approaches that use formal methods primarily to validate or filter LLM outputs, \emph{CodeLogician} uses LLMs to help construct explicit formal models (``mental models") of software systems, enabling automated reasoning to answer rich semantic questions beyond binary verification outcomes.
While the current implementation builds on \emph{ImandraX} and benefits from its unique features designed for large-scale industrial formal verification and software understanding, the framework is explicitly designed to be reasoner-agnostic and can support additional formal reasoning backends.

To rigorously evaluate mathematical reasoning about software logic, we introduce a new benchmark dataset \emph{code-logic-bench}\footnote{\url{https://github.com/imandra-ai/code-logic-bench}} targeting the previously unaddressed middle ground between theorem proving and software engineering benchmarks. 
The benchmark measures correctness and efficacy of reasoning about program state spaces, control flow, coverage constraints, decision boundaries, and edge cases, with ground truth defined via formal modeling and automated decomposition.

Using this benchmark set, we compare LLM-only reasoning against LLMs augmented with \emph{CodeLogician}. 
Across all evaluated models, formal augmentation with \emph{CodeLogician} yields substantial and consistent improvements, closing a 41--47 percentage point gap in reasoning accuracy and achieving complete coverage while LLM-only approaches remain approximate or incorrect. 
These results demonstrate that the neurosymbolic integration of LLMs with formal reasoning engines is essential for scaling program analysis beyond heuristic reasoning toward rigorous, autonomous software understanding and formal verification.
\end{abstract}

\maketitle
\newpage
\tableofcontents

\newpage
\section{Introduction}
The rapid adoption of Large Language Models (LLMs) in software development has brought the notion of ``reasoning about code'' into the mainstream.
LLMs are now widely used for tasks such as code completion, refactoring, documentation, and test generation, and have demonstrated strong capabilities in understanding control flow and translating between programming languages.
However, despite these advances, LLMs remain fundamentally limited in their ability to perform precise, exhaustive mathematical reasoning about program behavior.

In parallel, the field of automated reasoning---often referred to as symbolic AI or formal methods---has long focused on mathematically precise analysis of programs expressed as formal models.
Decades of research and industrial deployment have produced powerful techniques for verifying correctness, exhaustively exploring program behaviors, and identifying subtle edge cases.
Yet these techniques have historically required specialized expertise, limiting their accessibility and integration into everyday software development workflows.

This paper introduces \emph{CodeLogician}, a neuro-symbolic framework that bridges these two paradigms.
\emph{CodeLogician} combines LLM-driven agents with \emph{ImandraX}, an automated reasoning engine that has been successfully applied in financial markets, safety-critical systems, and government agencies.
With \emph{CodeLogician}, LLMs are used to translate source code into executable formal models and to interpret reasoning results, while automated reasoning engines perform exhaustive mathematical analysis of those models.

\subsection{Beyond Verification-as-Filtering}

Much of the existing work that combines LLMs with formal methods treats symbolic systems as external ``validators".
In these approaches, LLMs generate code, specifications, or proofs, and automated reasoning tools are applied post hoc to check consistency or correctness, typically yielding a binary yes/no outcome.
While such verification-as-filtering is valuable for enforcing correctness constraints, it does not fundamentally expand the reasoning capabilities of LLMs.

In particular, verification-as-filtering does not help LLMs construct accurate internal models of program behavior, reason about large or infinite state spaces, or answer rich semantic questions involving constraints, decision boundaries, and edge cases.
As a result, LLMs remain confined to approximate and heuristic reasoning, even when formal tools are present in the pipeline.

\emph{CodeLogician} adopts a different perspective.
Rather than using formal methods to police LLM outputs, we use LLMs to \emph{construct explicit formal mental models} of software systems expressed in mathematical logic.
Automated reasoning engines are then applied directly to these models to answer questions that go far beyond satisfiability or validity, including exhaustive behavioral coverage, precise boundary identification, and systematic test generation.
In this setting, LLMs act as translators and orchestrators, while mathematical reasoning is delegated to tools designed explicitly for that purpose.

\subsection{Neuro-Symbolic Complementarity}

The strengths and weaknesses of LLMs closely mirror those of symbolic reasoning systems.
LLMs excel at translation, abstraction, and working with incomplete information, but struggle with precise logical reasoning.
Symbolic systems, by contrast, require strict formalization and domain discipline, but excel at exhaustive and exact reasoning once a model is available.

\emph{CodeLogician} embraces this neuro-symbolic complementarity.
LLMs bridge the gap between informal source code and formal logic, while automated reasoning engines provide mathematical guarantees about program behavior.
This separation of concerns enables a scalable and principled approach to reasoning about real-world software systems.

\subsection{ImandraX and Reasoner-Agnostic Design}

At the core of \emph{CodeLogician} lies \emph{ImandraX}, an automated reasoning engine operating over the Imandra Modeling Language (IML), a pure functional language based on a subset of OCaml with extensions for specification and verification~\cite{passmoreImandraAutomatedReasoning2020}.
Beyond traditional theorem proving and formal verification, \emph{ImandraX} supports advanced analysis techniques such as region decomposition, enabling systematic exploration of program state spaces and automated generation of high-coverage test suites.

While \emph{CodeLogician} currently integrates \emph{ImandraX} as its primary reasoning backend, the framework is explicitly designed to be \emph{reasoner-agnostic}.
Additional formal reasoning tools can be incorporated without architectural changes, allowing CodeLogician to evolve alongside advances in automated reasoning.

\subsection{The Missing Benchmark: Mathematical Reasoning About Software}

Despite rapid progress in evaluating LLMs, existing benchmarks fall into two largely disconnected categories.
Mathematical proof benchmarks evaluate abstract symbolic reasoning over mathematical domains, but do not address reasoning about real-world software behavior.
Engineering benchmarks, such as those focused on debugging or patch generation, evaluate practical development skills but do not require precise or exhaustive semantic reasoning.

What is missing is a benchmark that evaluates an LLM's ability to apply \emph{mathematical reasoning to executable software logic}.
To address this gap, we introduce a new benchmark dataset \emph{code-logic-bench}~\cite{code_logic_bench} designed to measure reasoning about program state spaces, constraints, decision boundaries, and edge cases.
Ground truth is defined via formal modeling and automated reasoning, enabling objective evaluation of correctness, coverage, and boundary precision.
The benchmark supports principled comparison between LLM-only reasoning and LLMs augmented with formal reasoning via \emph{CodeLogician}.

\subsection{Contributions}

This paper makes the following contributions:
\begin{itemize}
    \item We present \emph{CodeLogician}, a neuro-symbolic framework that teaches LLM-driven agents to construct formal models of software systems and reason about them using automated reasoning engines.
    \item We introduce a new benchmark dataset targeting mathematical reasoning about software logic, filling a critical gap between existing mathematical and engineering-oriented benchmarks.
    \item We provide a rigorous empirical evaluation demonstrating substantial improvements when LLMs are augmented with formal reasoning, compared to LLM-only approaches.
\end{itemize}

\section{CodeLogician overview}
\label{sec:codelogicianoverview}

\begin{figure}
  \frame{\includegraphics[width=1\textwidth]{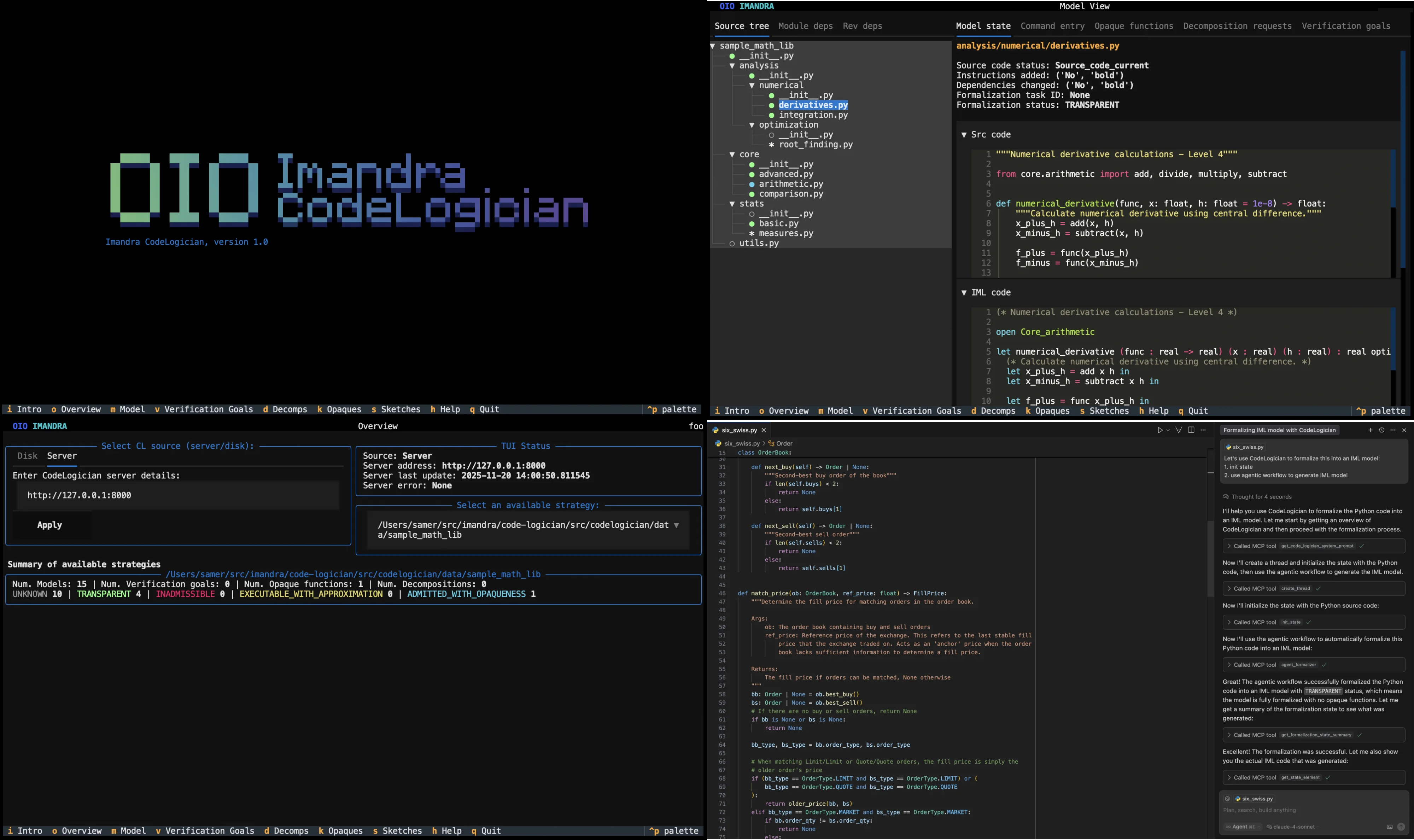}}
  \caption{\emph{Imandra CodeLogician}, neurosymbolic agent and framework for precise reasoning about software logic, with CLI, MCP and VS Code interfaces. Available from \url{www.codelogician.dev}.}\label{fig:codelogician}
\end{figure}

\emph{CodeLogician} is a neuro-symbolic, agentic governance framework for AI-assisted software development.
It provides a uniform orchestration layer that exposes formal reasoning engines (called "reasoners"), autoformalization agents, and analysis workflows through consistent interfaces. It supports multiple modes of use, including direct invocation of reasoners, agent-driven auto-formalization, and LLM-assisted utilities for documentation and analysis. It is available as both CLI and server-based workflows—with an interactive TUI that can invoke agents over entire directories—making formal reasoning, verification, and test-case generation accessible across development pipelines.

Rather than competing with LLM-based coding tools, \emph{CodeLogician} is designed to augment and guide them by supplying mathematically grounded feedback, proofs, counterexamples, and exhaustive behavioral analyses.

At its core, \emph{CodeLogician} serves as a source of \emph{mathematical ground truth}.
Where LLM-based tools operate probabilistically, \emph{CodeLogician} anchors reasoning about software behavior in formal logic, ensuring that conclusions are derived from explicit models and automated reasoning rather than statistical inference.

\subsection{Positioning and Scope}

\emph{CodeLogician} targets reasoning tasks that arise above the level of individual lines of code, where system behavior emerges from the interaction of components, requirements, and environments.
Typical application areas include architectural reasoning, multi-system integration, protocol and workflow validation, and functional correctness of critical business logic.

While \emph{CodeLogician} can, in principle, reason about low-level implementation details, its primary value lies in governing system-level behavior.
Domains such as memory safety, SQL injection detection, or hardware-adjacent concerns are already well served by specialized static analyzers and runtime tools.
\emph{CodeLogician} complements these point solutions by providing a framework for validating architectural intent, invariants, and cross-component interactions with mathematical completeness.

\subsection{Extensible Reasoner-Oriented Design}

\emph{CodeLogician} is explicitly designed as a multi-reasoner framework.
Its first release integrates the \emph{ImandraX} automated reasoning engine, which supports executable formal models, proof-oriented verification, and exhaustive state-space analysis.
However, the framework is intentionally reasoner-agnostic and is architected to host additional formal tools (e.g., temporal logic model checkers such as TLA+) under the same orchestration and governance layer.

This design allows \emph{CodeLogician} to evolve alongside advances in formal methods without coupling its user-facing workflows to a single solver, logic, or verification paradigm.

\subsection{Modes of Operation}

\emph{CodeLogician} exposes its capabilities through several complementary modes of operation, all backed by the same underlying orchestration and reasoning infrastructure.

\paragraph{Direct invocation.}
In direct mode, users or external tools invoke reasoning engines and agents explicitly.
This mode supports fine-grained control over formalization, verification, and analysis steps and is particularly useful for expert users and debugging complex models.

\paragraph{Agent-mediated workflows.}
In agentic modes, \emph{CodeLogician} coordinates LLM-driven agents that automatically formalize source code, refine models, generate verification goals, and invoke reasoning backends.
These workflows enable end-to-end analysis with minimal human intervention while retaining the ability to request feedback when ambiguities or inconsistencies arise.

\paragraph{Bring-Your-Own-Agent (BYOA).}
\emph{CodeLogician} can be used as a reasoning backend for externally hosted LLM agents.
In this mode, an LLM interacts with \emph{CodeLogician} through structured commands, receives formal feedback, and iteratively improves its own formalization behavior.
This enables tight integration with existing LLM-based coding assistants while preserving a clear separation between statistical generation and logical reasoning.

\paragraph{Server-based operation.}
For large-scale or continuous analysis, \emph{CodeLogician} can be deployed in a server mode that monitors codebases or projects and incrementally updates formal models and analyses in response to changes.
This supports use cases such as continuous verification, regression analysis, and long-running agentic workflows.

\subsection{Governance and Feedback}

A central contribution of \emph{CodeLogician} is its role as a governance layer for AI-assisted development.
Rather than merely accepting or rejecting LLM outputs, \emph{CodeLogician} provides structured, semantically rich feedback grounded in formal reasoning.
This includes:
\begin{itemize}
    \item proofs of correctness when properties hold,
    \item counterexamples when properties fail,
    \item explicit characterization of decision boundaries and edge cases,
    \item quantitative coverage information derived from exhaustive analysis.
\end{itemize}

This feedback can be consumed by humans, external agents, or LLMs themselves, enabling iterative improvement and learning.
In this sense, \emph{CodeLogician} does not merely analyze code—it shapes how AI systems reason about software over time.

\subsection{Relationship to the \emph{CodeLogician IML Agent}}

The \emph{CodeLogician IML Agent} described in Section~\ref{sec:codelogician-agent} is one concrete realization of the framework’s capabilities.
It implements an end-to-end autoformalization pipeline that translates source code into executable formal models and invokes \emph{ImandraX} for analysis.

More generally, the framework supports multiple agents, reasoning engines, and workflows operating over shared representations and governed by the same orchestration logic.
This separation between framework and agents allows \emph{CodeLogician} to scale across domains, reasoning techniques, and modes of interaction.

\subsection{Summary}

\emph{CodeLogician} is a unifying framework for integrating LLMs with formal reasoning engines in a principled, extensible manner.
By providing orchestration, governance, and structured feedback around automated reasoning, it enables AI-assisted software development to move beyond heuristic reasoning toward mathematically grounded analysis.
The framework establishes a firm foundation upon which multiple agents and reasoners can coexist, evolve, and collaborate in the service of rigorous software engineering.

\section{Overview of ImandraX}
\label{sec:imandrax}

\begin{figure}
  \frame{\includegraphics[width=1\textwidth]{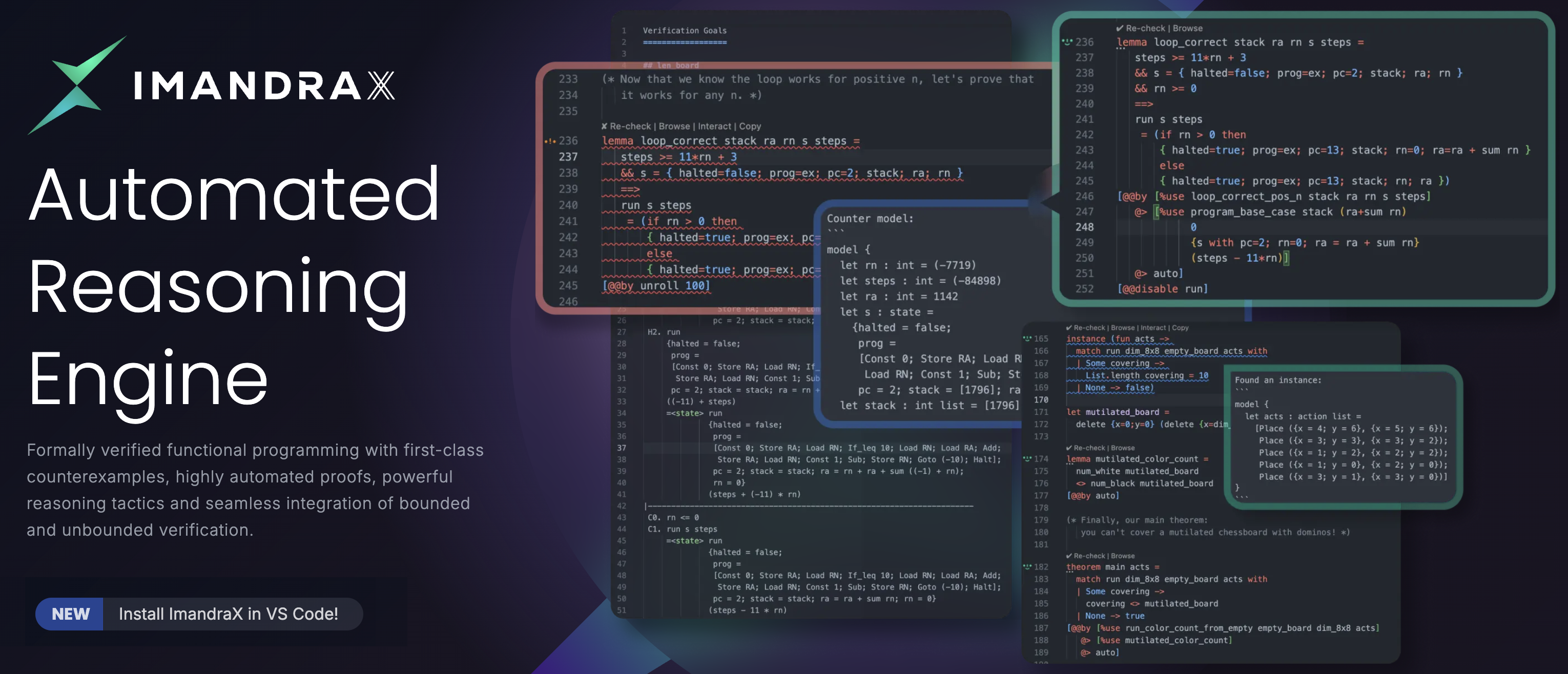}}
  \caption{The ImandraX automated reasoning engine and theorem prover. Interfaces are available for both humans (VS Code) and AI assistants (MCP and CLI), and may be installed from \url{www.imandra.ai/core}.}\label{fig:imandrax}
\end{figure}

\emph{ImandraX} (cf. Fig~\ref{fig:imandrax}) is an automated reasoning engine for analyzing executable mathematical models of software systems.
Its input language is the \emph{Imandra Modeling Language} (IML), a pure functional subset of OCaml augmented with directives for specification and reasoning (e.g., \texttt{verify}, \texttt{lemma} and \texttt{instance})~\cite{passmoreImandraAutomatedReasoning2020}.
IML models are ordinary programs, but they are also \emph{formal models}: \emph{ImandraX} contains a mechanized formal semantics for IML, and thus IML code is both executable code and rigorous mathematical logic to which \emph{ImandraX} applies automated reasoning.

\emph{CodeLogician} relies on \emph{ImandraX} as its primary reasoning backend~{\footnote{\url{https://www.imandra.ai/core}}}.
In the \emph{autoformalization} workflow (cf. Section~\ref{sec:autoformalization}), LLM-driven agents translate source code into executable IML models, make abstraction choices explicit, and introduce assumptions where required.
\emph{ImandraX} then performs the mathematical analysis of these models, enabling both proof-oriented verification and richer forms of behavioral understanding.

This section summarizes the two \emph{ImandraX} capabilities most heavily used by \emph{CodeLogician}: verification goals and state-space region decomposition.
As in Section~\ref{sec:autoformalization}, we emphasize that LLMs are used to \emph{construct} explicit formal mental models, while \emph{ImandraX} is used to \emph{reason} about them.

\subsection{Formal Models in the Imandra Modeling Language (IML)}
\label{sec:imandrax-iml}

IML inherits the modeling discipline discussed in Section~\ref{sec:autoformalization}:
models are pure (no side effects), statically typed functional programs in \emph{ImandraX}'s fragment of OCaml, structured to support inductive reasoning modulo decision procedures.
Imperative constructs such as loops are represented via recursion, and systems with mutable state are modeled as state machines in which the evolving state is carried explicitly through function inputs and outputs.

The question is therefore not whether a system can be modeled, but at what level of abstraction it should be modeled (Section~\ref{sec:autoformalization}).
\emph{ImandraX} provides analysis capabilities that remain meaningful across these abstraction levels, provided the model is executable and its assumptions are made explicit.

\subsection{Verification Goals}
\label{sec:imandrax-vgs}

Formal verification in \emph{ImandraX} is performed via the statement and analysis of \emph{verification goals} (VGs): boolean-valued IML functions that encode properties of the system being analyzed. VGs are (typically) implicitly universally quantified: when one \emph{proves} a theorem in \emph{ImandraX}, one establishes that a boolean-valued VG is {\textbf{true}} for all possible inputs.
Unlike testing, which evaluates finitely many concrete executions, VGs quantify over (typically) infinite input spaces and yield either:
(i) a proof that the property holds universally, or
(ii) a counterexample witness demonstrating failure.

As programs and datatypes may involve recursion over structures of arbitrary depth, these proofs may involve induction. A unique feature of \emph{ImandraX} is its seamless integration of bounded and unbounded verification. Every goal may be subjected to both bounded model checking and full inductive theorem proving. Bounded proofs are automatic based on \emph{ImandraX}'s \emph{recursive function unrolling} modulo decision procedures, and typically take place in a setting for which they are complete for counterexamples. \emph{ImandraX} contains deep automation for constructing inductive proofs, including a lifting of the Boyer-Moore inductive waterfall to \emph{ImandraX}'s typed, higher-order setting~\cite{passmoreImandraAutomatedReasoning2020}. Moreover, \emph{ImandraX} contains a rich tactic language for the interactive construction of proofs and counterexamples, and for more general system and state-space exploration.

\begin{figure}
  \frame{\includegraphics[width=1\textwidth]{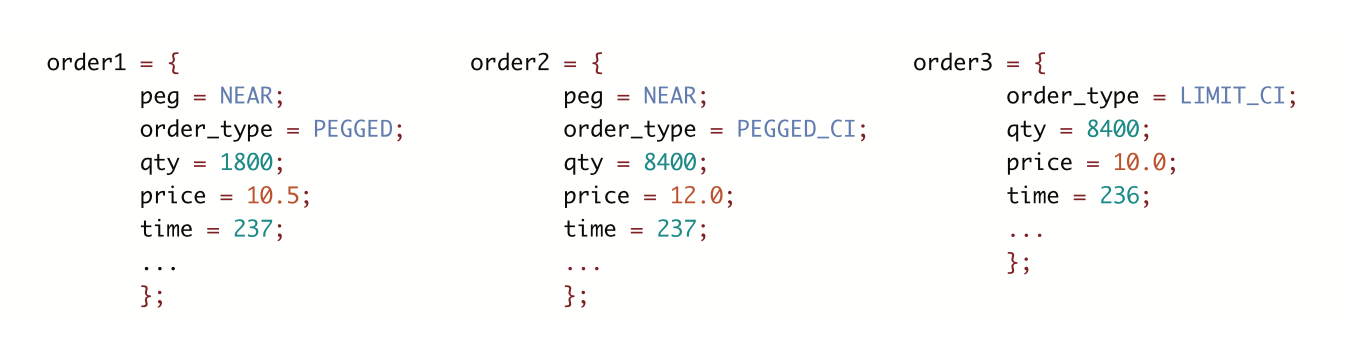}}
  \caption{Counterexample for Order Ranking Transitivity}\label{fig:ubs-cx}
\end{figure}

For example, consider the transitivity of an order ranking function for a trading venue~\cite{passmore-ignatovich-fvfa}. This may naturally be conjectured as a lemma, which may then be subjected to verification:

\begin{lstlisting}[language=ocaml]
lemma rank_transitivity side order1 order2 order3 mkt =
  order_higher_ranked(side,order1,order2,mkt) &&
  order_higher_ranked(side,order2,order3,mkt)
  ==>
  order_higher_ranked(side,order1,order3,mkt)
\end{lstlisting}

In practice, verification is iterative: counterexamples often reveal missing preconditions, modeling gaps, or specification errors. For example, in Fig.~\ref{fig:ubs-cx} we see a (truncated version of) an ImandraX-synthesized counterexample for {\textbf{rank\_transitivity}} of a real-world trading venue~\cite{passmore-ignatovich-fvfa}.

In \emph{ImandraX}, all counterexamples are ``first-class'' objects which may be reflected in the runtime and computed with directly via \emph{ImandraX}'s {\textbf{eval}} directive.
This unified computational environment for programs, conjectures, counterexamples and proofs is important for the analysis of real-world software.
Verification goals should immediately be amenable to automated analysis, counterexamples for false goals should be synthesized efficiently, and these counterexamples should be available for direct execution through the system under analysis, enabling rapid triage and understanding for goal violations and false conjectures.
This use of counterexamples aligns with the autoformalization perspective: constructing a good formal model is a refinement process in which assumptions and intended semantics are progressively made explicit, and counterexamples to conjectures can rapidly help one to fix faulty logic and recognize missing assumptions.

\subsection{Region Decomposition}
\label{sec:imandrax-decomp}

\begin{figure}[h!]
  \begin{center}
    \frame{\includegraphics[width=1\textwidth]{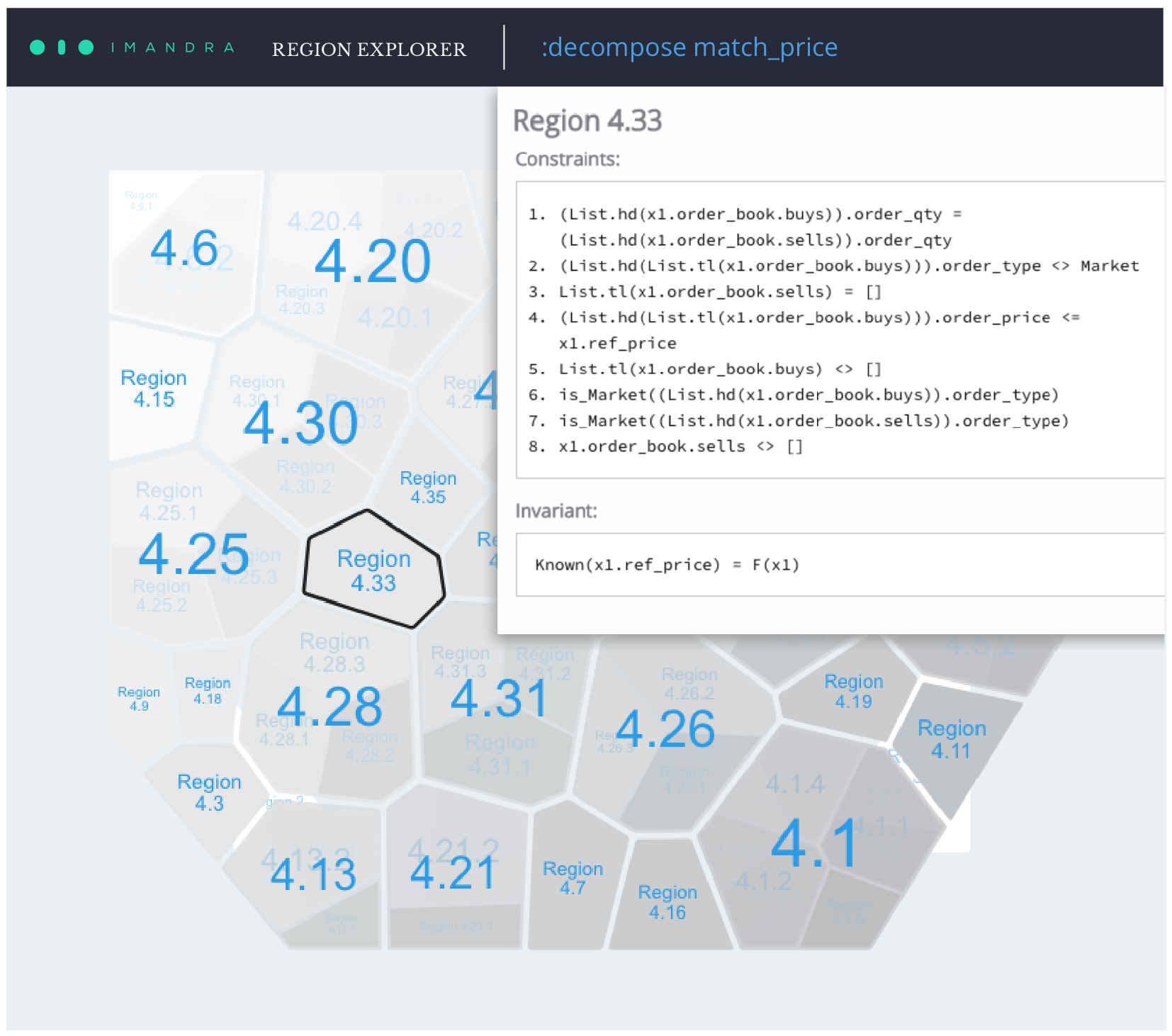}}
  \end{center}
  \caption{A principal region decomposition of a trading venue pricing function}\label{fig:voronoi}
\end{figure}

Many questions central to program understanding are not naturally expressed as binary VGs.
To support richer semantic analysis, \emph{ImandraX} provides \emph{region decomposition}, which partitions a function's input space into finitely many symbolic regions corresponding to distinct behaviors.

Each region comprises:
\begin{itemize}
  \item \emph{constraints} over inputs that characterize when execution enters the region,
  \item an \emph{invariant result} describing the output (or a symbolic characterization of it) for all inputs satisfying the constraints, and
  \item \emph{sample points} witnessing satisfiable instances of the constraints.
\end{itemize}

Region decomposition provides a mathematically precise notion of ``edge cases'' and directly supports the systematic test generation goals described in Section~\ref{sec:autoformalization}.

\begin{comment}
Consider the simple definitions of \lstinline|f| and \lstinline|g| shown in Fig.~\ref{fig:simple-code}:

\begin{figure}[t]
\begin{lstlisting}[language=ocaml]
let g x =
  if x > 22 then 9
  else 100 + x

let f x =
  if x > 99 then 100
  else if x < 70 && x > 23 then 89 + x
  else if x > 20 then g x + 20
  else if x > -2 then 103
  else 99
[@@decomp top ()]

\end{lstlisting}
\caption{Simple IML functions and a call to region decomposition}
\end{figure}\label{fig:simple-code}
\end{comment}

\emph{ImandraX} decomposes function into regions corresponding to the distinct path conditions induced by their control flow, each with explicit constraints and invariant outputs.
These decompositions may be performed relative to logical side-conditions and configurable abstraction boundaries called \textit{basis functions} to tailor state-space exploration for a particular goal.
In real software systems, small boundary differences (e.g., a flipped inequality) often lead to qualitatively different outcomes; region decomposition isolates these differences and makes the boundaries of logically invariant behavior explicit and auditable.

\subsection{Focusing Decomposition: Side Conditions and Basis}
\label{sec:imandrax-focus}

\begin{figure}[h!]
  \begin{center}
    \frame{\includegraphics[width=1\textwidth]{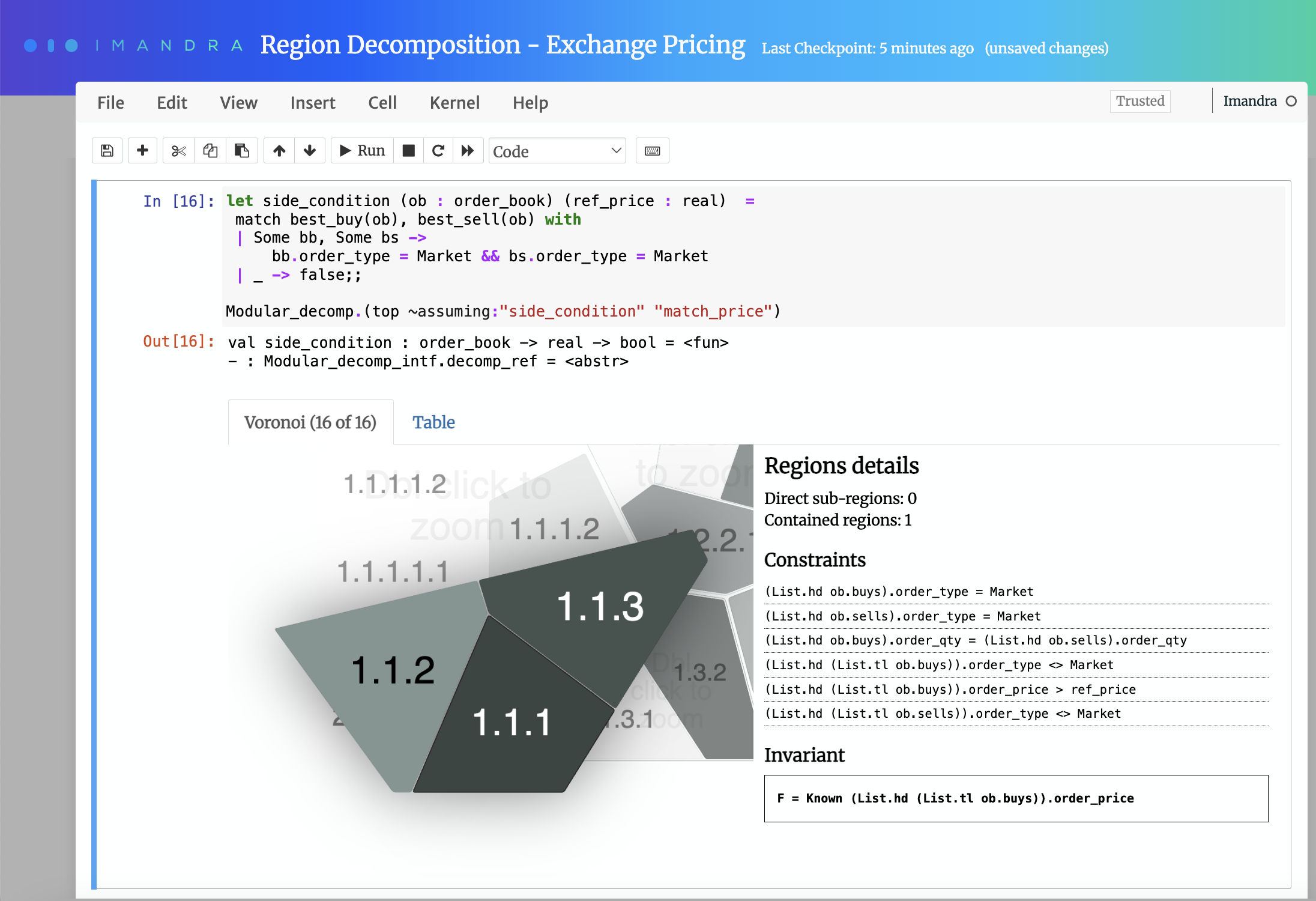}}
  \end{center}
  \caption{A refined decomposition of the venue pricing function with a side-condition\label{fig:voronoi-2}}
\end{figure}

Autoformalization often produces models whose full state spaces are too large---or simply too broad---for a given analysis objective.
\emph{ImandraX} provides two complementary mechanisms to focus decomposition, mirroring the abstraction discipline in Section~\ref{sec:autoformalization}.

\paragraph{Side conditions.}
A \emph{side condition} is a boolean predicate with the same signature as the decomposed function.
When supplied, \emph{ImandraX} explores only regions for which the side condition holds.
This is useful for restricting analysis to scenarios of operational or regulatory interest.
Figure~\ref{fig:voronoi-2} illustrates a region decomposition query augmented with a side-condition.

\paragraph{Basis functions.}
A \emph{basis} specifies functions to be treated as atomic during decomposition.
\emph{ImandraX} will not explore their internal branching structure, reducing the number of regions and improving interpretability.
Basis selection provides a principled way to ``abstract away'' auxiliary computations while retaining their influence on the surrounding model.

Together, side-conditions and basis functions provide a rich query language for targeting the analysis of system state spaces towards particular classes of behaviors~\cite{imandra-six-swiss-exchange-pricing,imandra-supervised-learning}. 

\subsection{Test Generation from Regions}
\label{sec:imandrax-tests}

Each region produced by decomposition includes sample points satisfying its constraints.
These witnesses can be extracted to generate tests that correspond directly to semantic distinctions in program behavior, rather than to ad hoc input distributions.
In \emph{CodeLogician}, this capability underpins automated test generation: agents use regions to produce tests that cover all behavioral cases discovered by decomposition, consistent with the ``exhaustive coverage'' objective motivating autoformalization.

\subsection{Summary}

\emph{ImandraX} provides the formal reasoning substrate for \emph{CodeLogician}.
Verification goals support proof-oriented analysis, while region decomposition exposes the structure of program state spaces, makes decision boundaries explicit, and enables systematic test-case generation.
Together with the autoformalization process (Section~\ref{sec:autoformalization}), these capabilities allow LLM-driven agents to construct formal mental models of software and to answer rich semantic questions beyond binary verification outcomes.

\section{Autoformalization of source code}
\label{sec:autoformalization}

Autoformalization is the process of translating executable source code into precise, mathematical models suitable for automated reasoning.
While this translation is partially automated by CodeLogician, it relies on a particular way of thinking about software---one that emphasizes explicit state, clear abstraction boundaries, and well-defined assumptions.
Just as developers strive to write \emph{clear code}, effective automated reasoning depends on writing or synthesizing \emph{clear formal models}.

\subsection{Pure Functional Modeling}

All formal models produced by CodeLogician are expressed in the Imandra Modeling Language (IML), a pure functional subset of OCaml.
Purity here means the absence of side effects: functions cannot mutate global state or influence values outside their scope.
Instead, all state changes must be made explicit via function arguments and return values.

This restriction is not a fundamental limitation.
Programs with side effects can be systematically transformed into state-machine models in which the evolving program state is explicitly carried through computations.
Such techniques are well established in the literature and trace back to foundational work by Boyer and Moore~\cite{boyer-moore-acl,clinc-stack}.
For example, this (infinite) state-machine model approach underlies successful formal verification of highly complex industrial systems, including large-scale financial market infrastructure~\cite{passmore-ignatovich-fvfa,passmoreLessonsLearnedIndustrialization2021a}, complex hardware models~\cite{wintersteiger-floats-2025}, robotic controllers integrating deep neural networks~\cite{desmartin_et_al:imandra-nn-proof-checker}, and high-frequency communication protocols~\cite{imandra-ipl}.

\subsection{Key Modeling Constructs}

Several structural features of IML are particularly important for automated reasoning:

\paragraph{Static Types}
IML is a statically typed functional language.
All variable types are determined before execution and cannot change dynamically.
This eliminates entire classes of ambiguity present in dynamically typed languages.
During autoformalization, CodeLogician extracts type information from source programs---using static annotations where available and inference otherwise---and enforces type consistency in the generated models.
Type errors detected by ImandraX often correspond directly to logical inconsistencies in the original program.

\paragraph{Recursion and Induction}
Loops in imperative source code are rewritten as recursive functions.
Both \texttt{for} and \texttt{while} constructs are mapped to structurally recursive definitions.
This transformation is essential because reasoning about unbounded iteration requires inductive reasoning, which is naturally expressed over recursive functions.

\paragraph{State Machines}
Many real-world programs are best modeled as state machines.
These machines are typically infinite-state, as the state may include complex data structures such as lists, maps, or trees in addition to numeric values.
State-machine modeling is particularly effective for representing code with side effects and for capturing the behavior of complex reactive systems, such as trading platforms or protocol handlers.

\subsection{Abstraction Levels}

A central question in autoformalization is not merely whether a program can be modeled, but \emph{at what level of abstraction it should be modeled}.
The appropriate abstraction depends on the properties one wishes to analyze.

We distinguish three broad abstraction levels:

\paragraph{High-Level Models}
High-level models typically arise from domain-specific languages (DSLs) designed to describe complex systems concisely.
In such cases, direct use of CodeLogician is often unnecessary.
Instead, the DSL itself is translated into IML.
For example, the Imandra Protocol Language (IPL) provides a compact notation for symbolic state-transition systems and compiles into IML with significant structural expansion.
Reasoning at this level focuses on system-wide invariants and high-level behavioral properties.

\paragraph{Mid-Level Models}
Application-level software represents the primary target for CodeLogician.
At this level, source programs can be translated directly into IML without introducing an intermediate DSL.
This category includes most business logic, data-processing pipelines, and control-heavy application code.

\paragraph{Low-Level Models}
Low-level code, such as instruction sequences or hardware-near implementations, can also be modeled in ImandraX, but doing so requires an explicit formalization of the execution environment (e.g., memory model, virtual machine, or instruction semantics).
Such models are outside the intended scope of CodeLogician, which assumes access to semantic abstractions of the underlying execution platform.

\subsection{Dealing with Unknowns}

Real-world programs inevitably rely on external libraries, services, or system calls.
Autoformalization therefore requires explicit handling of unknown or partially specified behavior.
CodeLogician adopts techniques analogous to mocking in software testing, but within a formal reasoning framework.

\subsubsection{Opaque Functions}

IML supports \emph{opaque functions}, which declare a function’s type without specifying its implementation.
Opaque functions allow models to type-check and support reasoning without making any assumptions about the function’s internal behavior.

For example, calls to external libraries such as pseudo-random number generators are introduced as opaque functions during autoformalization.
Reasoning about code that depends on such functions remains sound, but necessarily conservative.

\subsubsection{Axioms and Assumptions}

To refine reasoning about opaque functions, users (or agents) may introduce axioms that constrain their behavior.
These axioms encode assumptions about external components, such as value ranges or monotonicity properties.
Introducing axioms narrows the set of possible behaviors considered during reasoning and can significantly strengthen the conclusions that can be drawn.

Importantly, axioms are explicit and local: reasoning results are valid only under the stated assumptions, making the modeling process auditable and transparent.

\subsubsection{Approximation Functions}

In many cases, opaque functions can be replaced with sound approximations.
For common mathematical functions, CodeLogician provides libraries of approximation functions, such as bounded Taylor-series expansions for trigonometric operations.
These approximations allow models to remain executable and enable test generation, while still supporting rigorous reasoning within known bounds.

\subsection{Summary}

Autoformalization is not merely a syntactic translation process.
It is a disciplined approach to modeling software behavior in which state, control flow, assumptions, and abstraction boundaries are made explicit.
By combining LLM-driven translation with principled formal modeling techniques, CodeLogician enables automated reasoning engines to answer deep semantic questions about software behavior that are inaccessible to LLMs alone.

\section{CodeLogician IML Agent}
\label{sec:codelogician-agent}

CodeLogician is a neuro-symbolic agent framework designed to enable Large Language Models (LLMs) to reason rigorously about software systems.
Rather than asking LLMs to reason directly about source code, CodeLogician teaches LLM-driven agents to construct explicit formal models in mathematical logic and to delegate semantic reasoning to automated reasoning engines.

The current implementation integrates the ImandraX automated reasoning engine together with static analysis tools and auxiliary knowledge sources within an agentic architecture built on LangGraph.
CodeLogician exposes both low-level programmatic commands and high-level agentic workflows, allowing users and external agents to flexibly combine automated reasoning with guided interaction.
When required information is missing or inconsistencies are detected, the agent can request human feedback, enabling mixed-initiative refinement.
The framework can also be embedded into external agent systems via a remote graph API.

\subsection{From Source Code to Formal Models}

At the core of CodeLogician lies the process of \emph{autoformalization} (Section~\ref{sec:autoformalization}).
Given a source program written in a conventional programming language (e.g., Python), the agent incrementally constructs an executable formal model in the Imandra Modeling Language (IML).

Autoformalization is not a single translation step, but a structured refinement process that involves:
\begin{itemize}
    \item extracting structural, control-flow, and type information from the source program,
    \item refactoring imperative constructs into pure functional representations,
    \item making implicit assumptions explicit (e.g., about external libraries or partial functions),
    \item synthesizing an IML model that is admissible to the reasoning engine.
\end{itemize}

The resulting model is functionally equivalent to the source program at the chosen level of abstraction and suitable for mathematical analysis by ImandraX.

\subsection{Verification Goals and Logical Properties}

Once a formal model has been constructed, CodeLogician assists in formulating \emph{verification goals} (VGs).
A verification goal is a boolean-valued predicate over the model that expresses a property expected to hold for all possible inputs.
Unlike traditional testing, verification goals quantify symbolically over infinite input spaces and admit mathematical proof or counterexample.

For example, consider a ranking function used in a financial trading system.
A fundamental correctness requirement is transitivity: if one order ranks above a second, and the second ranks above a third, then the first must rank above the third.
Such properties are expressed directly as verification goals in IML and discharged by ImandraX.

When a verification goal does not hold, ImandraX produces a counterexample witness.
These counterexamples are concrete executions that violate the property and often expose subtle logical flaws that are difficult to detect through testing or informal inspection.
In industrial case studies, this approach has uncovered violations of core invariants in production systems that would otherwise remain hidden.

\subsection{Beyond Yes/No Verification}

While the proof or refutation of a verification goal determines whether or not a property is \emph{true}, many forms of program understanding require richer, non-binary kinds of semantic information.
CodeLogician therefore relies not only on formal verification, but also on exhaustive behavioral analysis techniques provided by ImandraX.

In particular, CodeLogician uses state-space exploration to identify decision boundaries, semantic edge cases, and invariant behaviors.
These capabilities enable explanations, documentation, and test generation that go far beyond yes/no verification and are central to CodeLogician’s value proposition.

\subsection{Agent Architecture and State}

CodeLogician is implemented as a stateful agent whose execution is organized around an explicit \emph{agent state}.
This state captures all artifacts relevant to the reasoning process and evolves as the agent performs analysis and refinement.

The agent state includes:
\begin{itemize}
    \item the original source program,
    \item extracted formalization metadata (types, control flow, assumptions),
    \item the generated IML model,
    \item the model’s admission and executability status,
    \item verification goals, decomposition requests, and related artifacts.
\end{itemize}

Agent operations fall into three broad categories:
\begin{itemize}
    \item \emph{State transformations}, which update the agent state explicitly (e.g., modifying the source program or model);
    \item \emph{Agentic workflows}, which orchestrate multi-step processes such as end-to-end autoformalization and analysis;
    \item \emph{Low-level commands}, which perform individual actions such as model admission, verification, or decomposition.
\end{itemize}

This separation allows users and external agents to mix automated and guided reasoning as appropriate.

\subsection{Model Refinement and Executability}

Autoformalization proceeds as an iterative refinement process.
The immediate objective is to produce a model that is \emph{admissible} to ImandraX, meaning that it is well-typed and free of missing definitions.
A further objective is to make the model \emph{executable}, eliminating opaque functions either by implementation or by sound approximation.

These two stages correspond to:
\begin{itemize}
    \item \emph{Admitted models}, which can be verified but may contain unresolved abstractions;
    \item \emph{Executable models}, which support full behavioral analysis and test generation.
\end{itemize}

This refinement process mirrors established practices in model-based software engineering and digital twinning.
By working with an executable mathematical model, CodeLogician enables rigorous reasoning about systems whose direct analysis in the source language would be impractical.

\subsection{Assessing Model Correctness}

A natural question arises when constructing formal models: how do we know the model is correct?
CodeLogician adopts a notion of correctness based on \emph{functional equivalence} at the chosen abstraction level.
Two programs are considered equivalent if they produce the same observable outputs for the same inputs within the modeled domain.

Exhaustive behavioral analysis plays a central role in this assessment.
By enumerating all distinct behavioral regions of the model, CodeLogician provides a structured and auditable view of program behavior.
This makes mismatches between intended and modeled semantics explicit and actionable.

\subsection{Region decomposition, test case generation, and refinement}

The final stage is where Imandra shows its power. We perform region decomposition on the IML code to obtain a finite number of symbolically described regions such that (a) the union of all regions covers the entire state-space, and (b) in each region, the behavior of the system is invariant.

Consider the following IML code that models order discount logic:
\begin{lstlisting}[language=ocaml]
type priority = Standard | Premium
type order = { amount: int; customer: priority }

let discount = fun o ->
  match o.customer with
  | Premium -> if o.amount > 100 then 20 else 10
  | Standard -> if o.amount > 100 then 10 else 0
\end{lstlisting}

Region decomposition partitions the input space into four distinct regions based on customer type and order amount. For each region, ImandraX provides concrete witness values that satisfy the region's constraints.
To bridge the gap between these symbolic results and executable test cases, we developed \href{https://pypi.org/project/imandrax-codegen/}{\texttt{imandrax-codegen}}, an open-source code generation tool. Given the decomposition artifacts, the tool automatically generates code including type definitions and test cases. We use Python as the target language in this example; support for additional languages is planned:
\begin{lstlisting}[language=python]
@dataclass
class order:
    amount: int
    customer: priority

@dataclass
class Standard:
    pass

@dataclass
class Premium:
    pass

priority = Standard | Premium

def test_1():
    """test_1

    - invariant: 0
    - constraints:
        - not (o.customer = Premium)
        - o.amount <= 100
    """
    result: int = discount(o=order(0, Standard()))
    expected: int = 0
    assert result == expected

def test_2():
    """test_2

    - invariant: 10
    - constraints:
        - not (o.customer = Premium)
        - o.amount >= 101
    """
    result: int = discount(o=order(101, Standard()))
    expected: int = 10
    assert result == expected

# ... test_3, test_4 are omitted for brevity
\end{lstlisting}

This workflow ensures that we have test cases that cover all possible behavioral regions of the system---in this example, all four combinations of customer priority and order amount threshold.

The entire pipeline is configurable, allowing for a balance between automation and human intervention.
For example, the user can choose to double-check refactored code in intermediate steps, provide additional human feedback on error messages or let the agent handle it by itself, etc. And as mentioned in the previous section, FDB provides critical RAG support in all stages of the pipeline.

\subsection{Interfaces}

CodeLogician exposes both input and output interfaces designed to be simultaneously user-friendly and programmatically accessible.
The interfaces follow a simple input schema paired with a rich, structured output format, enabling use in interactive development environments as well as integration into larger agent-based systems.

The input interface accepts source programs (currently Python) as textual input, together with optional configuration parameters that control the autoformalization and reasoning process.
This minimal input design allows CodeLogician to be invoked uniformly from command-line tools, IDEs, or external agents.

The output interface is structured using Pydantic models, providing strong validation, typing, and serialization guarantees.
The output schema captures detailed artifacts produced during each reasoning attempt, including:
\begin{itemize}
    \item generated IML models,
    \item compilation and admission results,
    \item verification outcomes and counterexamples,
    \item region decomposition results and associated metadata.
\end{itemize}
This structured representation makes CodeLogician suitable both for standalone use and as a component within larger neuro-symbolic pipelines.

CodeLogician also exposes configuration parameters that control the degree of automation and human intervention.
Parameters such as \lstinline|confirm_refactor|, \lstinline|decomp|, \lstinline|max_attempts|, and\\
\lstinline|max_attempts_wo_human_feedback| allow users to trade off full automation against mixed-initiative interaction, depending on the complexity and criticality of the target code.

\subsubsection{Programmatic Access via \lstinline|python|{RemoteGraph}}

CodeLogician can be directly integrated with other LangGraph agents using the \lstinline|RemoteGraph| interface.
By specifying the agent’s URI within Imandra Universe and supplying an API key, external agents can invoke CodeLogician as a remote reasoning component.
This enables composition with other agentic workflows, including multi-agent systems and tool-augmented LLM pipelines.
In addition to programmatic access, CodeLogician is also available through a web-based chat interface in Imandra Universe.

\subsubsection{VS Code Extension}

CodeLogician is distributed with a free Visual Studio Code (VS Code) extension~\cite{extension}, providing a lightweight and accessible interface for interactive use.
The extension communicates with the CodeLogician agent via the \lstinline|RemoteGraph| interface and exposes core commands through the VS Code command palette.

The extension is built around a command-queue abstraction that allows users to sequence multiple operations—such as autoformalization, verification, and decomposition—before execution.
It supports simultaneous formalization of multiple source files and provides visibility into the agent’s internal state, including generated models and reasoning artifacts.

\begin{figure}[htbp]
    \centering
    \includegraphics[width=\textwidth]{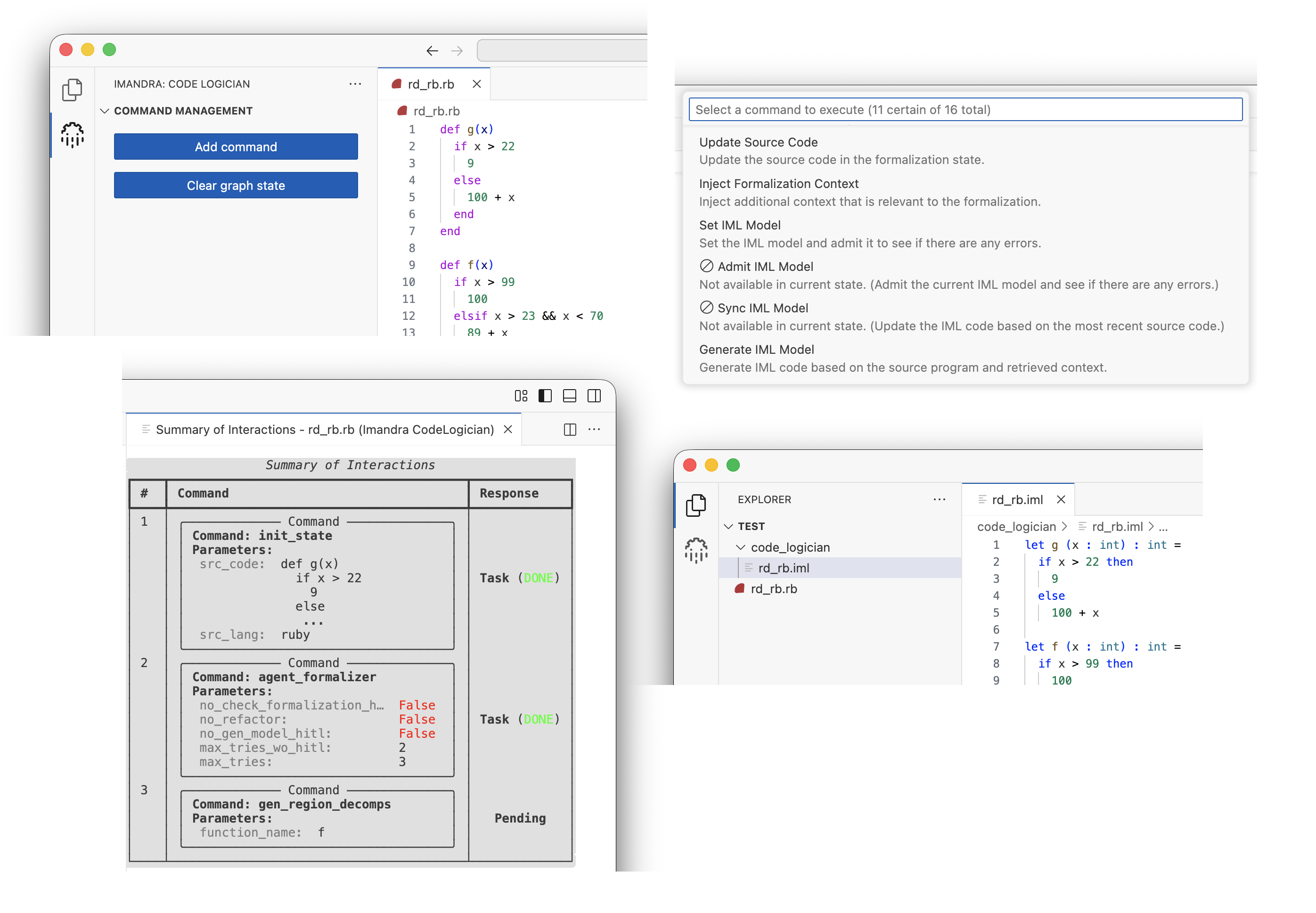}
    \caption{CodeLogician VS Code extension, showing command management, agent state visualization, and a generated IML model.}
    \label{fig:vscode}
\end{figure}

Figure~\ref{fig:vscode} illustrates the VS Code extension’s command management system, state viewer, and an example IML model produced by CodeLogician.

\subsection{Summary}

The CodeLogician IML agent operationalizes the neuro-symbolic approach advocated in this paper.
LLMs are used to construct formal mental models of software systems, while automated reasoning engines provide mathematically precise analysis.
By combining autoformalization, verification goals, and exhaustive behavioral analysis within an agentic framework, CodeLogician enables rigorous reasoning about real-world software systems beyond the capabilities of LLM-only approaches.

\section{CodeLogician Server}
% Failsafe: define the node style locally (redefining is harmless)
\tikzset{
  modulebox/.style={
    rectangle, draw, rounded corners,
    minimum width=3.5cm, minimum height=1cm,
    align=center
  }
}

\label{sec:server}

The \emph{CodeLogician} server provides a persistent backend for project-wide formalization and analysis.
It manages formalization tasks, maintains analysis state across sessions, and serves multiple client interfaces (TUI and CLI).
Unlike one-shot invocation modes, the server is designed for \emph{continuous} and \emph{incremental} operation: it caches results, reacts to code changes, and recomputes only the minimal set of affected artifacts required to keep formal models consistent with the underlying code base.

\subsection{Server Responsibilities}

At a high level, the server is responsible for:
\begin{itemize}
    \item \textbf{Persistent analysis state}: storing formalization artifacts and their status across sessions;
    \item \textbf{Project management}: tracking multiple repositories, configurations, and formalization strategies;
    \item \textbf{Incremental updates}: monitoring file and dependency changes and triggering minimal re-formalization;
    \item \textbf{Caching}: avoiding redundant computation by reusing results for unchanged modules;
    \item \textbf{Interface multiplexing}: serving requests from interactive and programmatic clients via a shared backend.
\end{itemize}

These responsibilities enable a workflow in which developers and agents can repeatedly query and refine formal models while the server maintains a coherent, up-to-date view of the formalized state of the project.

\subsection{Strategies: Project-Wide Formalization via PyIML}
\label{sec:pyiml-strategy}

The server executes \emph{strategies}, configureable pipelines that formalize a code base and manage dependencies between its components.
For Python projects, \emph{CodeLogician} provides the \emph{PyIML strategy}, which translates Python source code into IML models suitable for reasoning in \emph{ImandraX}.

Given a project directory, the PyIML strategy:
\begin{enumerate}
    \item \textbf{Analyzes} Python modules and infers their dependency structure from imports;
    \item \textbf{Translates} source constructs (functions, classes, types) into corresponding IML artifacts;
    \item \textbf{Manages} formalization across the codebase in dependency-consistent order;
    \item \textbf{Maintains} a project \emph{metamodel} capturing modules, dependencies, and formalization status.
\end{enumerate}

The strategy also handles \emph{opaque functions} (Section~\ref{sec:autoformalization}) by preserving type information while abstracting implementation details when direct translation is not possible.
This allows the system to admit models and begin formal reasoning even in the presence of external libraries or currently-unknown semantics, while making the loss of executability explicit.

\subsection{Metamodel-Based Formalization}
\label{sec:metamodel-formalization}

Formalizing large software systems requires reasoning not only about individual source files, but also about their interdependencies.
To support scalable project-wide analysis, the \emph{CodeLogician} server maintains an explicit \emph{metamodel} that captures the structure, dependencies, and formalization status of a collection of models corresponding to a repository.

\subsection{Core Definitions}

\paragraph{Model}
A \emph{model} represents the formalization state of a single source file or module.
Each model contains:
\begin{itemize}
  \item \texttt{src\_code}: the source code of the corresponding file or module;
  \item \texttt{formalization\_status}: the current level of formalization achieved;
  \item \texttt{dependencies}: references to other models on which this model depends;
  \item \texttt{dependency\_context}: an aggregated formal representation (in IML) of available dependencies, used during formalization.
\end{itemize}

\paragraph{Metamodel}
A \emph{metamodel} is a container for all models associated with a given project directory (repository).
It provides operations for model insertion, deletion, update, and dependency management, and serves as the authoritative representation of the project’s formalization state.

\paragraph{Interdependence}
By interdependence, we mean that models import symbols (e.g., functions, classes, constants) from other project files or third-party libraries.
By default, a model can be formalized even when referenced symbols lack formal implementations; however, such symbols must be treated as \emph{opaque}, which limits the strength of reasoning that can be performed.

\paragraph{Autoformalization}
Autoformalization is the process by which the \emph{CodeLogician} agent translates source code, together with its dependency context, into an IML model and submits it to the reasoning engine.
This process may produce artifacts such as verification goals, counterexamples, and test cases.

\subsection{Formalization Levels}

Each model is assigned a formalization level reflecting how much semantic information is available:
\begin{itemize}
  \item \textbf{Unknown}: no formalization has been performed;
  \item \textbf{Error during validation}: admission failed due to issues such as inconsistent types, missing symbols, or non-termination;
  \item \textbf{Admitted with opaqueness}: the model has been admitted but contains opaque symbols;
  \item \textbf{Admitted transparent}: the model has been admitted and contains no opaque elements.
\end{itemize}

The \texttt{dependency\_context} is computed by aggregating all available IML models of dependencies and injecting this context alongside the model’s own source code during formalization.

\paragraph{Objective.}
The objective of the server is to maximize the number of models that are formalized, and to achieve the highest possible level of formalization for each model, while ensuring that the most up-to-date source code, dependency context, and human input are reflected.

\subsection{Conditions Triggering Re-Formalization}

A model requires re-formalization when:
\begin{itemize}
  \item its underlying source code has changed or has never been formalized; or
  \item the source code or formalization status of one or more dependencies (direct or indirect) has changed.
\end{itemize}

\subsection{Autoformalization Workflow}

\begin{figure}[htbp]
  \centering
  \includegraphics[scale=0.15]{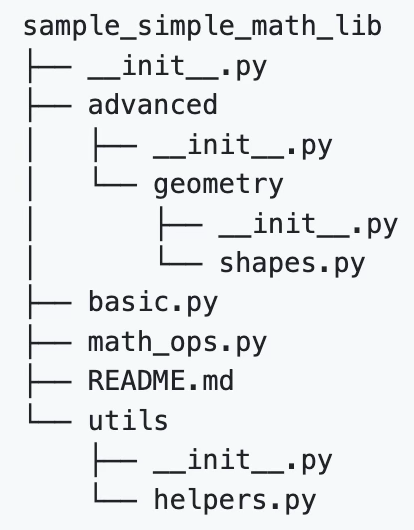}
  \caption{Example project structure.}
  \label{fig:server-project-outline}
\end{figure}

\paragraph{Initial state.}
Figure~\ref{fig:server-project-outline} shows an example repository with a hierarchical structure in which modules import symbols from one another.

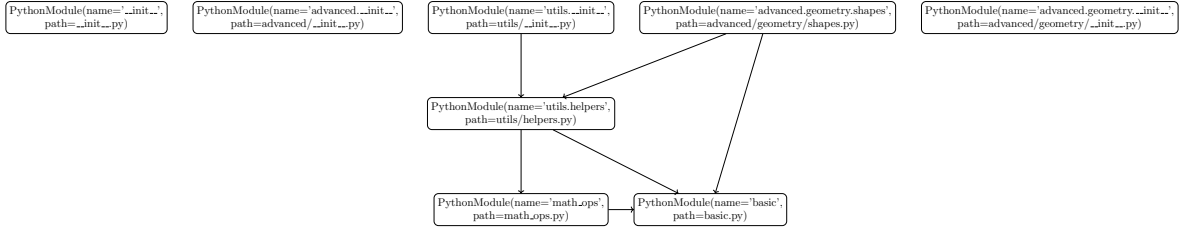
\begin{figure}[htbp]
  \centering
  \begin{adjustbox}{max width=\textwidth}

  \begin{tikzpicture}[node distance=1.5cm and 0.8cm]
    % Top row nodes
    \node[box] (init) {PythonModule(name='\_\_init\_\_',\\ path=\_\_init\_\_.py)};
    \node[box, right=of init] (adv) {PythonModule(name='advanced.\_\_init\_\_',\\ path=advanced/\_\_init\_\_.py)};
    \node[box, right=of adv] (utils) {PythonModule(name='utils.\_\_init\_\_',\\ path=utils/\_\_init\_\_.py)};
    \node[box, right=of utils] (shapes) {PythonModule(name='advanced.geometry.shapes',\\ path=advanced/geometry/shapes.py)};
    \node[box, right=of shapes] (geom) {PythonModule(name='advanced.geometry.\_\_init\_\_',\\ path=advanced/geometry/\_\_init\_\_.py)};

    % Second row
    \node[box, below=2cm of utils] (helpers) {PythonModule(name='utils.helpers',\\ path=utils/helpers.py)};

    % Third row
    \node[box, below=2cm of helpers] (mathops) {PythonModule(name='math\_ops',\\ path=math\_ops.py)};

    % Fourth row
    \node[box, right=of mathops] (basic) {PythonModule(name='basic',\\ path=basic.py)};

    % Arrows
    \draw[->] (utils) -- (helpers);
    \draw[->] (shapes) -- (helpers);
    \draw[->] (helpers) -- (mathops);
    \draw[->] (mathops) -- (basic);
    \draw[->] (helpers) -- (basic);
    \draw[->] (shapes) -- (basic);
    \end{tikzpicture}
    \end{adjustbox}
  \caption{Dependency graph inferred from imports.}
  \label{fig:server-dep-graph}
\end{figure}

The dependency graph inferred from imports is shown in Figure~\ref{fig:server-dep-graph}.
An edge from a source node to a target node indicates that the source imports symbols from the target.

\begin{figure}[htbp]
  \centering
  \includegraphics[width=\textwidth]{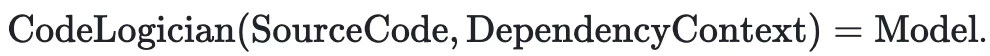}
  \caption{Topological formalization order.}
  \label{fig:server-topo-sequence}
\end{figure}

Autoformalization proceeds bottom-up along a topological ordering of the dependency graph (Figure~\ref{fig:server-topo-sequence}).

\begin{figure}[htbp]
  \centering
  \includegraphics[width=\textwidth]{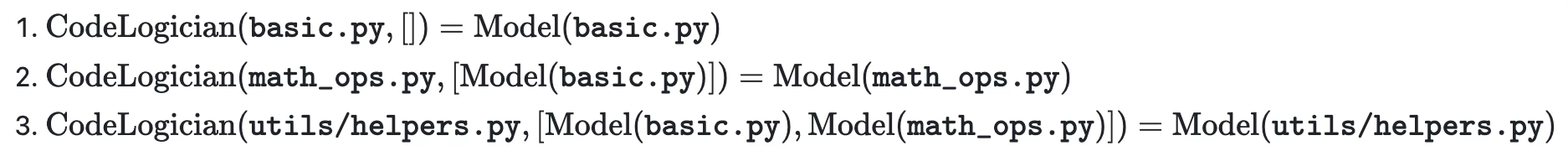}
  \caption{Formalization sequence for a target module.}
  \label{fig:server-topo-execution}
\end{figure}

Suppose we wish to reason about functions in \texttt{utils/helpers.py}.
To do so, CodeLogician formalizes that module and all of its dependencies in dependency order (Figure~\ref{fig:server-topo-execution}).

\begin{figure}[htbp]
  \centering
  
  \begin{subfigure}{\linewidth}
  \begin{adjustbox}{max width=\textwidth}
  \begin{tikzpicture}[node distance=1.5cm and 0.8cm]
    % Top row nodes
    \node[box] (init) {PythonModule(name='\_\_init\_\_',\\ path=\_\_init\_\_.py)};
    \node[box, right=of init] (adv) {PythonModule(name='advanced.\_\_init\_\_',\\ path=advanced/\_\_init\_\_.py)};
    \node[box, right=of adv] (utils) {PythonModule(name='utils.\_\_init\_\_',\\ path=utils/\_\_init\_\_.py)};
    \node[box, right=of utils] (shapes) {PythonModule(name='advanced.geometry.shapes',\\ path=advanced/geometry/shapes.py)};
    \node[box, right=of shapes] (geom) {PythonModule(name='advanced.geometry.\_\_init\_\_',\\ path=advanced/geometry/\_\_init\_\_.py)};

    % Second row - HIGHLIGHTED
    \node[box, fill=green, below=2cm of utils] (helpers) {PythonModule(name='utils.helpers',\\ path=utils/helpers.py)};

    % Third row - HIGHLIGHTED
    \node[box, fill=green, below=2cm of helpers] (mathops) {PythonModule(name='math\_ops',\\ path=math\_ops.py)};

    % Fourth row - HIGHLIGHTED
    \node[box, fill=green, right=of mathops] (basic) {PythonModule(name='basic',\\ path=basic.py)};

    % Arrows
    \draw[->] (utils) -- (helpers);
    \draw[->] (shapes) -- (helpers);
    \draw[->] (helpers) -- (mathops);
    \draw[->] (mathops) -- (basic);
    \draw[->] (helpers) -- (basic);
    \draw[->] (shapes) -- (basic);
    \end{tikzpicture}
    \end{adjustbox}
    \end{subfigure}
    \\[10pt]
    \begin{subfigure}{\linewidth}
    \begin{adjustbox}{max width=\textwidth}
    \begin{tabular}{|l|l|l|l|c|c|}
    \hline
    \textbf{Path} & \textbf{Model Name} & \textbf{Formalization Status} & \textbf{Sync Status} & \textbf{Can Formalize} & \textbf{Needs Formalization} \\
    \hline
    basic.py & basic.iml & transparent & synced & \ding{51}  & \ding{55} \\
    \hline
    math\_ops.py & math\_ops.iml & transparent & source\_modified & \ding{51} & \ding{55} \\
    \hline
    utils/helpers.py & helpers.iml & transparent & dependency\_changed & \ding{51} & \ding{55} \\
    \hline
    advanced/geometry/init.py & init.iml & unknown & never\_formalized & \ding{51} & \ding{51} \\
    \hline
    advanced/geometry/shapes.py & shapes.iml & unknown & never\_formalized & \ding{51} & \ding{51} \\
    \hline
    utils/\_\_init\_\_.py & init.iml & unknown & never\_formalized & \ding{51} & \ding{51} \\
    \hline
    advanced/\_\_init\_\_.py & init.iml & unknown & never\_formalized & \ding{51} & \ding{51} \\
    \hline
    \_\_init\_\_.py & init.iml & unknown & never\_formalized & \ding{51} & \ding{51} \\
    \hline
    \end{tabular}
    \end{adjustbox}
    \end{subfigure}

  \caption{Summary of formalization status, formalized modules highlighted.}
  \label{fig:server-formalized}
\end{figure}
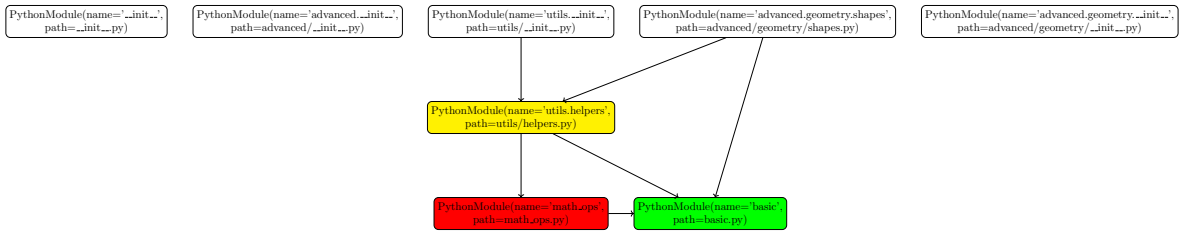

After completion, formalized models appear highlighted, while unformalized modules remain unchanged (Figures~\ref{fig:server-formalized}. % and~\ref{fig:server-summary}).

% \begin{figure}[htbp]
%   \centering
%   % \includegraphics[width=\textwidth]{images/server/5-summary-frm-status.png}
%   \caption{Summary of formalization status.}
%   \label{fig:server-summary}
% \end{figure}

\subsection{Minimal Re-Formalization Strategy}

When source code changes, the server computes precise diffs to identify the minimal set of models that require re-formalization, ensuring consistency while avoiding unnecessary recomputation.

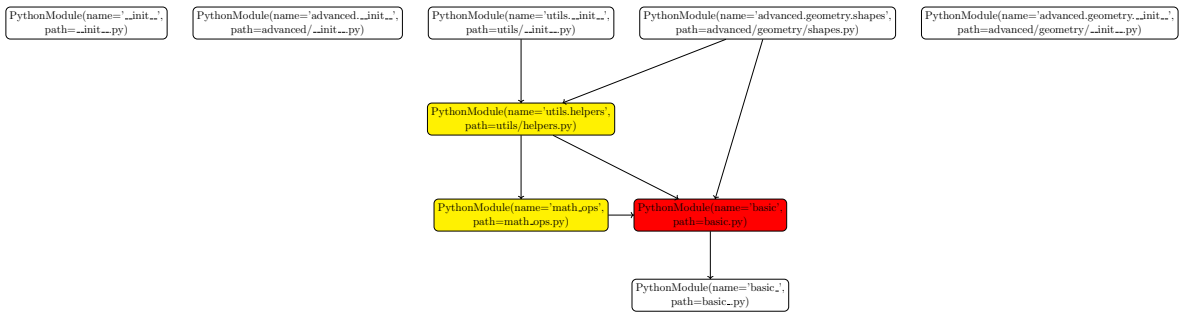
\begin{figure}[htbp]
  \centering
    
    \begin{subfigure}{\linewidth}
    \begin{adjustbox}{max width=\textwidth}
    \begin{tikzpicture}[node distance=1.5cm and 0.8cm]
    % Top row nodes
    \node[box] (init) {PythonModule(name='\_\_init\_\_',\\ path=\_\_init\_\_.py)};
    \node[box, right=of init] (adv) {PythonModule(name='advanced.\_\_init\_\_',\\ path=advanced/\_\_init\_\_.py)};
    \node[box, right=of adv] (utils) {PythonModule(name='utils.\_\_init\_\_',\\ path=utils/\_\_init\_\_.py)};
    \node[box, right=of utils] (shapes) {PythonModule(name='advanced.geometry.shapes',\\ path=advanced/geometry/shapes.py)};
    \node[box, right=of shapes] (geom) {PythonModule(name='advanced.geometry.\_\_init\_\_',\\ path=advanced/geometry/\_\_init\_\_.py)};

    % Second row - YELLOW
    \node[box, fill=yellow, below=2cm of utils] (helpers) {PythonModule(name='utils.helpers',\\ path=utils/helpers.py)};

    % Third row - RED
    \node[box, fill=red, below=2cm of helpers] (mathops) {PythonModule(name='math\_ops',\\ path=math\_ops.py)};

    % Fourth row - GREEN
    \node[box, fill=green, right=of mathops] (basic) {PythonModule(name='basic',\\ path=basic.py)};

    % Arrows
    \draw[->] (utils) -- (helpers);
    \draw[->] (shapes) -- (helpers);
    \draw[->] (helpers) -- (mathops);
    \draw[->] (mathops) -- (basic);
    \draw[->] (helpers) -- (basic);
    \draw[->] (shapes) -- (basic);
    \end{tikzpicture}
    \end{adjustbox}
    \end{subfigure}
    \\[10pt]
    \begin{subfigure}{\linewidth}
    \begin{adjustbox}{max width=\textwidth}
    \begin{tabular}{|l|l|l|l|c|c|}
    \hline
    \textbf{Path} & \textbf{Model Name} & \textbf{Formalization Status} & \textbf{Sync Status} & \textbf{Can Formalize} & \textbf{Needs Formalization} \\
    \hline
    basic.py & basic.iml & transparent & synced & \ding{51}  & \ding{55} \\
    \hline
    math\_ops.py & math\_ops.iml & transparent & source\_modified & \ding{51} & \ding{51} \\
    \hline
    utils/helpers.py & helpers.iml & transparent & dependency\_changed & \ding{55} & \ding{51} \\
    \hline
    advanced/geometry/init.py & init.iml & unknown & never\_formalized & \ding{51} & \ding{51} \\
    \hline
    advanced/geometry/shapes.py & shapes.iml & unknown & never\_formalized & \ding{55} & \ding{51} \\
    \hline
    utils/\_\_init\_\_.py & init.iml & unknown & never\_formalized & \ding{55} & \ding{51} \\
    \hline
    advanced/\_\_init\_\_.py & init.iml & unknown & never\_formalized & \ding{51} & \ding{51} \\
    \hline
    \_\_init\_\_.py & init.iml & unknown & never\_formalized & \ding{51} & \ding{51} \\
    \hline
    \end{tabular}
    \end{adjustbox}
  \end{subfigure}
  \caption{Source modification without dependency changes.}
  \label{fig:server-case1}
\end{figure}

\paragraph{Case 1: Source modified, dependencies unchanged.}
If \texttt{math\_ops.py} is modified without altering dependency relations (Figure~\ref{fig:server-case1}), the modified module is marked as \emph{source modified}, and only downstream formalized dependencies are updated.

\paragraph{Case 2: New dependency added.}
If a new module is introduced (Figure~\ref{fig:server-case2}), formalization proceeds in dependency order to restore consistency.

\begin{figure}[htbp]
  \centering

  \begin{subfigure}{\linewidth}
  \begin{adjustbox}{max width=\textwidth}
  \begin{tikzpicture}[node distance=1.5cm and 0.8cm]
    % Top row nodes
    \node[box] (init) {PythonModule(name='\_\_init\_\_',\\ path=\_\_init\_\_.py)};
    \node[box, right=of init] (adv) {PythonModule(name='advanced.\_\_init\_\_',\\ path=advanced/\_\_init\_\_.py)};
    \node[box, right=of adv] (utils) {PythonModule(name='utils.\_\_init\_\_',\\ path=utils/\_\_init\_\_.py)};
    \node[box, right=of utils] (shapes) {PythonModule(name='advanced.geometry.shapes',\\ path=advanced/geometry/shapes.py)};
    \node[box, right=of shapes] (geom) {PythonModule(name='advanced.geometry.\_\_init\_\_',\\ path=advanced/geometry/\_\_init\_\_.py)};

    % Second row - YELLOW
    \node[box, fill=yellow, below=2cm of utils] (helpers) {PythonModule(name='utils.helpers',\\ path=utils/helpers.py)};

    % Third row - YELLOW
    \node[box, fill=yellow, below=2cm of helpers] (mathops) {PythonModule(name='math\_ops',\\ path=math\_ops.py)};

    % Fourth row - RED
    \node[box, fill=red, right=of mathops] (basic) {PythonModule(name='basic',\\ path=basic.py)};

    % Fifth row - NEW NODE
    \node[box, below=1.5cm of basic] (basic2) {PythonModule(name='basic\_',\\ path=basic\_.py)};

    % Arrows
    \draw[->] (utils) -- (helpers);
    \draw[->] (shapes) -- (helpers);
    \draw[->] (helpers) -- (mathops);
    \draw[->] (mathops) -- (basic);
    \draw[->] (helpers) -- (basic);
    \draw[->] (shapes) -- (basic);
    \draw[->] (basic) -- (basic2);
    \end{tikzpicture}
    \end{adjustbox}
    \end{subfigure}
    \\[10pt]
  \begin{subfigure}{\linewidth}
  \begin{adjustbox}{max width=\textwidth}
  \begin{tabular}{|l|l|l|l|c|c|}
    \hline
    \textbf{Path} & \textbf{Model Name} & \textbf{Formalization Status} & \textbf{Sync Status} & \textbf{Can Formalize} & \textbf{Needs Formalization} \\
    \hline
    basic\_.py & basic\_.iml & unknown & never\_formalized & \ding{51} & \ding{51} \\
    \hline
    basic.py & basic.iml & transparent & source\_modified & \ding{55} & \ding{51} \\
    \hline
    math\_ops.py & math\_ops.iml & transparent & dependency\_changed & \ding{55} & \ding{51} \\
    \hline
    utils/helpers.py & helpers.iml & transparent & dependency\_changed & \ding{55} & \ding{51} \\
    \hline
    advanced/geometry/init.py & init.iml & unknown & never\_formalized & \ding{51} & \ding{51} \\
    \hline
    advanced/geometry/shapes.py & shapes.iml & unknown & never\_formalized & \ding{55} & \ding{51} \\
    \hline
    utils/\_\_init\_\_.py & init.iml & unknown & never\_formalized & \ding{55} & \ding{51} \\
    \hline
    advanced/\_\_init\_\_.py & init.iml & unknown & never\_formalized & \ding{51} & \ding{51} \\
    \hline
    \_\_init\_\_.py & init.iml & unknown & never\_formalized & \ding{51} & \ding{51} \\
    \hline
    \end{tabular}
    \end{adjustbox}
    \end{subfigure}
  \caption{Addition of a new dependency}
  \label{fig:server-case2}
\end{figure}

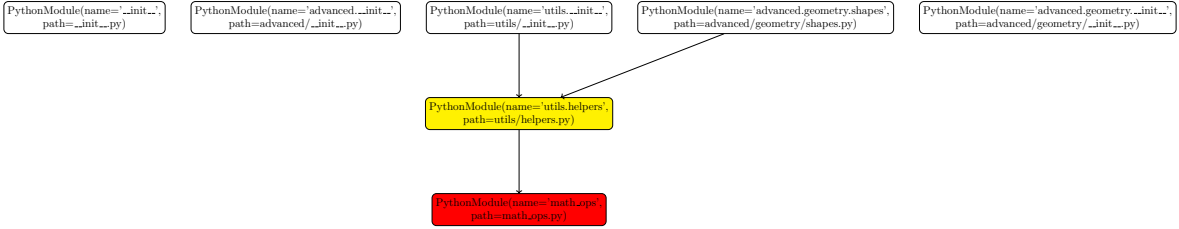
\begin{figure}[htbp]
  \centering
  \begin{adjustbox}{max width=\textwidth}
  \begin{tikzpicture}[node distance=1.5cm and 0.8cm]
    % Top row nodes
    \node[box] (init) {PythonModule(name='\_\_init\_\_',\\ path=\_\_init\_\_.py)};
    \node[box, right=of init] (adv) {PythonModule(name='advanced.\_\_init\_\_',\\ path=advanced/\_\_init\_\_.py)};
    \node[box, right=of adv] (utils) {PythonModule(name='utils.\_\_init\_\_',\\ path=utils/\_\_init\_\_.py)};
    \node[box, right=of utils] (shapes) {PythonModule(name='advanced.geometry.shapes',\\ path=advanced/geometry/shapes.py)};
    \node[box, right=of shapes] (geom) {PythonModule(name='advanced.geometry.\_\_init\_\_',\\ path=advanced/geometry/\_\_init\_\_.py)};

    % Second row - YELLOW
    \node[box, fill=yellow, below=2cm of utils] (helpers) {PythonModule(name='utils.helpers',\\ path=utils/helpers.py)};

    % Third row - RED
    \node[box, fill=red, below=2cm of helpers] (mathops) {PythonModule(name='math\_ops',\\ path=math\_ops.py)};

    % Arrows
    \draw[->] (utils) -- (helpers);
    \draw[->] (shapes) -- (helpers);
    \draw[->] (helpers) -- (mathops);
    \end{tikzpicture}
    \end{adjustbox}
  \caption{Merged dependencies.}
  \label{fig:server-case3}
\end{figure}

\paragraph{Case 3: Dependencies merged.}
When multiple modules are merged, the server recomputes formalization only for affected nodes (Figure~\ref{fig:server-case3}).

\begin{figure}[htbp]
  \centering
  \begin{adjustbox}{max width=\textwidth}
  \begin{tikzpicture}[node distance=1.5cm and 0.8cm]
  
    % Top row nodes
    \node[box] (init) {PythonModule(name='\_\_init\_\_',\\ path=\_\_init\_\_.py)};
    \node[box, right=of init] (adv) {PythonModule(name='advanced.\_\_init\_\_',\\ path=advanced/\_\_init\_\_.py)};
    \node[box, right=of adv] (utils) {PythonModule(name='utils.\_\_init\_\_',\\ path=utils/\_\_init\_\_.py)};
    \node[box, right=of utils] (shapes) {PythonModule(name='advanced.geometry.shapes',\\ path=advanced/geometry/shapes.py)};
    \node[box, right=of shapes] (geom) {PythonModule(name='advanced.geometry.\_\_init\_\_',\\ path=advanced/geometry/\_\_init\_\_.py)};

    % Second row - GREEN
    \node[box, fill=green, below=2cm of utils] (helpers) {PythonModule(name='utils.helpers',\\ path=utils/helpers.py)};

    % Third row - GREEN
    \node[box, fill=green, below=2cm of helpers] (mathops) {PythonModule(name='math\_ops',\\ path=math\_ops.py)};

    % Fourth row - GREEN
    \node[box, fill=green, right=of mathops] (basic) {PythonModule(name='basic',\\ path=basic.py)};

    % Arrows
    \draw[->] (utils) -- (helpers);
    \draw[->] (helpers) -- (mathops);
    \draw[->] (mathops) -- (basic);
    \draw[->] (helpers) -- (basic);
    \draw[->] (shapes) -- (basic);
    \end{tikzpicture}
    \end{adjustbox}
  \caption{Removal of an upstream dependency.}
  \label{fig:server-case4}
\end{figure}
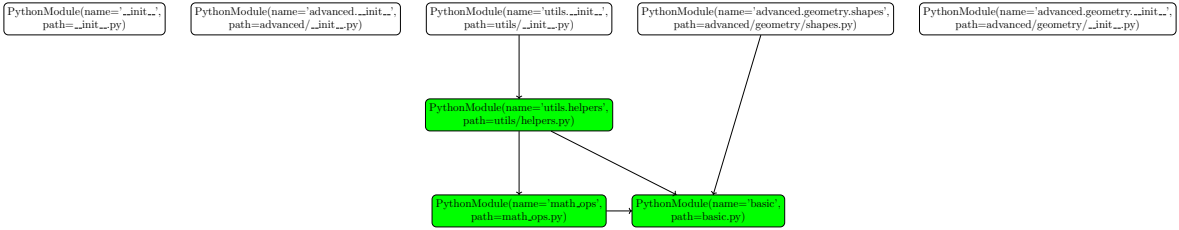

\paragraph{Case 4: Upstream dependency removed.}
If a dependency is removed from an unformalized module, no re-formalization is required for already formalized nodes (Figure~\ref{fig:server-case4}).

\subsection{Summary and Error Resilience}

In response to file updates, \emph{CodeLogician}:
\begin{enumerate}
  \item updates the dependency graph;
  \item computes the minimal set of required re-formalization tasks;
  \item avoids unnecessary re-formalization by targeting only affected models.
\end{enumerate}

To ensure robustness during active development, the server uses graceful degradation.
If parsing or dependency errors occur, it preserves the last known-good dependency graph and formal models, tracks errors for debugging, and automatically recovers once inconsistencies are resolved.

\section{Performance}
\subsection{Dataset}
We have created 50 models that have the following characteristics:
\begin{itemize}
  \item \textbf{State Machine Architecture} \\
  Each model defines explicit states and manages transitions between those states via step functions.

  \item \textbf{Complex Multi-Entity Systems} \\
  All models represent systems with multiple interacting components, for example:
  \begin{itemize}
    \item BGP path vector routing: speakers, neighbors, routes
    \item NPC pathfinding: characters, obstacles, terrain
    \item Smart warehouse auto-controller: robots, inventory, orders, zones
    \item Options market maker: options, positions, hedges
    \item Nuclear reactor safety system: control rods, safety systems, parameters
    \item TLS handshake protocol: client, server, certificates, sessions
  \end{itemize}

  \item \textbf{Temporal Evolution} \\
  Models explicitly track state changes over time using counters and/or timestamps, and evolve via simulation step functions.

  \item \textbf{Decision Logic} \\
  Each model implements complex decision-making based on current state and parameters (e.g.\ routing decisions, pathfinding, task assignment, hedging, safety checks, protocol steps).
  This logic often exhibits characteristic interaction patterns, including:
  \begin{itemize}
    \item \textbf{Override patterns}, where special conditions trump normal logic except in specific sub-cases
    \item \textbf{Dead-zone patterns}, where integer boundary combinations create gaps leading to undefined behavior or unexpected fallback states
  \end{itemize}

  \item \textbf{Safety and Correctness Properties} \\
  Many models include explicit invariants or constraints that must always hold, such as safety limits, risk limits, or convergence conditions.
\end{itemize}

All models are rooted in real-world systems that exist in production environments. Each model includes three questions that probe different aspects of state machine understanding, such as enumerating distinct behavioral scenarios, identifying conditions under which specific behaviors occur, and determining whether safety or correctness properties can be violated.

We evaluate two approaches to answering these questions:
\begin{itemize}
  \item \textbf{LLM-only:} The LLM analyzes the Python source code directly and answers the questions based on its own reasoning over the code.

  \item \textbf{\emph{CodeLogician} (LLM + \emph{ImandraX}):} The model is translated into IML (Imandra Modeling Language), and ImandraX performs formal region decomposition and verification. The LLM then synthesizes answers from these formal analysis results.
\end{itemize}

\subsection{Metrics}
We design 7 metrics to evaluate the performance of LLM-only reasoning compared to \emph{CodeLogician}. Each metric is
designed to assess a specific aspect of reasoning about complex state machines and decision
logic.

\subsubsection{State Space Estimation Accuracy}
This metric measures how accurately the LLM estimates the number of distinct behavioral
scenarios or regions in the state space. The ground truth comes from the exact region count
produced by CodeLogician's decomposition analysis.
\[
\textbf{Formula:}\quad
\text{score} = \frac{1}{1 + \log_2(\text{diff} + 1)},
\quad \text{where} \quad
\text{diff} = \lvert n_{\text{llm}} - n_{\text{decomp}} \rvert
\]

Design rationale: The logarithmic penalty reflects that being off by orders of magnitude (e.g.,
estimating 10 scenarios when there are 1000) is fundamentally more problematic than small
absolute differences. The formula yields: perfect match (diff=0) → score=1.0, off by 1 →
score=0.5, off by 3 → score=0.33, and converges to 0 as the difference grows. When the LLM
provides only qualitative descriptions without any quantification, we assign score=0.0.
For extraction, we use the following rules: (a) explicit numbers are used directly, (b) ranges use
the midpoint, (c) approximations extract the implicit number, (d) qualitative descriptions with
implicit combinatorial reasoning (e.g., "4 × 5 × 3 combinations") are computed, and (e) pure
qualitative statements with no extractable quantity are marked as unknown.

\subsubsection{Outcome Precision}
This metric evaluates the precision of constraints and conditions in the LLM's answer compared
to qualitative descriptions. It measures whether the LLM identifies exact constraints that
determine system behavior.

\paragraph{Scoring Rubric:}
\begin{itemize}
  \item \textbf{1.0}: Exact match (both value and operator are correct).
  \item \textbf{0.8}: Value is correct, but the operator is approximately equivalent (e.g., \(\geq 20\) vs.\ \(> 19\)).
  \item \textbf{0.5}: Value is approximately correct within a reasonable range (e.g., \(\geq 20\) vs.\ \(\geq 18\)).
  \item \textbf{0.2}: Qualitative but directionally correct (e.g., ``high'' when the correct constraint is \(> 620\)).
  \item \textbf{0.0}: Incorrect or missing.
\end{itemize}

Design rationale: This metric distinguishes between responses that capture exact conditions
versus vague qualitative descriptions. The graduated scoring recognizes that approximate
answers are more useful than purely qualitative ones, even if not perfectly precise.

\subsubsection{Direction Accuracy}
This metric assesses whether the LLM reaches conclusions in the correct direction, particularly
for yes/no questions about system properties or behaviors. For example, does the LLM
correctly identify that certain scenarios can violate a safety property, or does it incorrectly claim
the property always holds?

\paragraph{Scoring Rubric:}
\begin{itemize}
  \item \textbf{1.0}: Completely correct conclusion with sound reasoning.
  \item \textbf{0.75}: Correct conclusion but with flawed or incomplete reasoning.
  \item \textbf{0.5}: Partially correct.
  \item \textbf{0.25}: Incorrect conclusion but identifies some relevant factors.
  \item \textbf{0.0}: Completely incorrect.
\end{itemize}

Design rationale: Getting the fundamental direction of an answer wrong (e.g., claiming a system
is safe when it has critical vulnerabilities) is a severe error in code understanding. This metric
prioritizes correctness of the conclusion while also rewarding sound reasoning processes.

\subsubsection{Coverage Completeness}
This metric measures the proportion of actual decision scenarios or behavioral regions that the
LLM identifies in its analysis. It evaluates how thoroughly the LLM explores the state space.

\paragraph{Scoring Approach (Highest Applicable Method).}
\begin{enumerate}[label=(\alph*)]
  \item If the LLM provides explicit scenario counts, compare them directly to the decomposition count.
  \item If the LLM enumerates specific scenarios, count the number of scenarios enumerated.
  \item If the LLM categorizes outcomes, compute the score as
    \[
      \frac{\text{\# categories identified}}{\text{\# categories in decomposition}} .
    \]
  \item If there is only a qualitative acknowledgment of multiplicity, assign a score of \(0.25\).
  \item If there is no meaningful coverage, assign a score of \(0.0\).
\end{enumerate}

Design rationale: Complete coverage is essential for sound verification. Missing scenarios can
hide critical bugs or edge cases. We use multiple scoring methods to accommodate different
response styles, always choosing the method most favorable to the LLM to avoid penalizing
valid reasoning approaches.

\subsubsection{Control Flow Understanding}
This metric assesses whether the LLM correctly captures the structure of control flow, including
branches, guards, overrides, and short-circuit behavior. We evaluate the following four aspects:

\begin{itemize}
  \item \textbf{Precedence of conditions}: Correct handling of logical precedence (e.g., \texttt{AND} before \texttt{OR}, explicit priority ordering).
  \item \textbf{Branching structure}: Correct identification of control structures such as \texttt{if--else--if} chains versus independent conditional checks.
  \item \textbf{Short-circuit behavior}: Recognition of early returns, guard clauses, or other forms of short-circuit evaluation.
  \item \textbf{Override logic}: Identification of safety or exception-handling logic that supersedes normal operation.
\end{itemize}

\paragraph{Scoring Formula.}
The score is computed as:
\[
  \text{score} = \frac{\#\text{correct} + 0.5 \times \#\text{partial}}{4} .
\]

\paragraph{Design Rationale.}
Control flow structure is fundamental to program behavior. Misunderstanding whether conditions are
evaluated sequentially (with short-circuiting) versus independently, or misinterpreting logical
precedence, leads to incorrect analysis. Each aspect is therefore assessed independently and
aggregated into a single score, with partial credit awarded to reflect nuanced but incomplete
understanding.

\subsubsection{Edge Case Detection}
This metric evaluates how many rare or unexpected edge cases the LLM identifies in its analysis.
An \emph{edge case} is defined as any behavioral region exhibiting at least one of the following
properties:

\begin{itemize}
  \item \textbf{Complex conjunctive conditions}: Three or more conjunctive constraints combined in a single condition.
  \item \textbf{Exact boundary conditions}: Equality constraints at precise thresholds (e.g., \(x = 10\) rather than \(x > 10\)).
  \item \textbf{Negation of the common case}: Scenarios such as sensor failure, timeout, or other exceptional modes.
  \item \textbf{Counterintuitive outcomes}: Behaviors that contradict naive or intuitive expectations.
  \item \textbf{Safety-critical violations}: Invariant violations, error states, or other safety-relevant failures.
\end{itemize}

\paragraph{Scoring Formula.}
The score is computed as:
\[
  \text{score} =
  \frac{\#\text{edge cases identified by the LLM}}
       {\#\text{total edge cases in the decomposition}} .
\]

\paragraph{Design Rationale.}
Edge cases are where defects most commonly occur. Real-world systems tend to fail at boundary
conditions, under rare combinations of constraints, or in situations that violate designer
assumptions. This metric therefore rewards systematic identification of such cases. The definition
is intentionally broad in order to capture multiple forms of non-obvious or unexpected behavior.

\subsubsection{Decision Boundary Clarity}
\paragraph{Decision Boundary Identification Metric.}
This metric measures whether the LLM identifies the \emph{exact constraints} that determine
decision boundaries in the system. Each region produced by decomposition is defined by Boolean
constraints—conditions on inputs such as:

\begin{itemize}
  \item \( \text{deviation} \geq 80\,\text{cm} \),
  \item \( \text{state\_of\_charge} < 20\% \),
  \item \( \text{connection\_count} \geq 3 \),
  \item \( \text{temperature} > 100^\circ\text{C} \).
\end{itemize}

These constraints may take the form of inequalities, equalities, or logical combinations thereof.

\paragraph{Scoring Formula.}
The score is computed as:
\[
  \text{score} =
  \frac{\#\text{constraints identified by the LLM}}
       {\#\text{constraints in the decomposition}} .
\]

\paragraph{Design Rationale.}
Decision boundaries represent critical points in the state space at which system behavior changes
qualitatively. Accurately identifying these constraints is essential for precise reasoning about
when different behaviors occur. Qualitative descriptions such as ``low battery'' or ``high
temperature'' are insufficient for formal analysis and are therefore penalized by this metric.

\paragraph{Applicability Note.}
If a metric is not applicable or irrelevant to a specific question (e.g., decision boundaries are
requested for a system that has none), that metric is excluded from the evaluation for that
question.

\paragraph{Overall Evaluation Context.}
Collectively, these seven metrics capture complementary dimensions of how LLM
deviates from CodeLogician's formally-grounded analysis. Together, they provide a multifaceted assessment of the value added by
formal methods tools. A perfect score of \(1.0\) across all metrics would indicate that an LLM can
achieve the same level of rigor and completeness as formal analysis without assistance. The gaps
revealed by these metrics quantify the specific ways in which formal decomposition and verification
extend beyond unaided code reading and natural-language reasoning.

\subsection{Evaluation}

Our evaluation proceeds in two stages: answer generation and answer evaluation.

\subsubsection{Answer Generation}
We obtain answers from five frontier models: GPT-5.2, Grok Code Fast 1, Claude Sonnet 4.5, Claude Opus 4.5, and Gemini 3 Pro.
For the LLM-only condition, each model analyzes the Python source code directly and answers the three questions per model based on its own reasoning.
For the CodeLogician condition, answers are derived from ImandraX's formal region decomposition and verification results, which provide exact scenario counts, precise constraint boundaries, and proof outcomes (proven or refuted with counterexamples).
The CodeLogician answers serve as ground truth for evaluation.

\subsubsection{Answer Evaluation (LLM-as-a-Judge)}
We employ the \emph{LLM-as-a-judge} methodology to evaluate how well LLM-only answers align with \emph{CodeLogician} answers.
Four evaluator models—Gemini 2.5 Flash, Claude Haiku 4.5, GPT-4o Mini, and Grok 4 Fast—independently score each LLM response according to the seven metric definitions, providing detailed reasoning for each score.
We aggregate these evaluations by averaging across the four evaluators to reduce bias from any single judge.
While LLM-based evaluation introduces some variability, we plan to complement these results with deterministic methods such as test case generation from region decomposition result in the future.

All data, code, and scripts to reproduce these results are publicly available at \url{https://github.com/imandra-ai/code-logic-bench}.

\subsection{Results}
\subsubsection{Overall Performance}

\paragraph{State Space Estimation Accuracy (0.186).}
This metric reveals the most dramatic performance gap.
The low scores reflect both severe underestimation—when LLMs attempt quantification but miss combinatorial explosion by orders of magnitude—and complete inability to quantify, where models resort to vague statements such as ``many scenarios'' without attempting enumeration.
This failure to grasp state space size has direct consequences: it leads to underestimation of verification effort, inadequate test budgets, and false confidence that manual analysis has covered sufficient ground.
In safety-critical systems, such miscalibration can mean the difference between rigorous validation and deploying systems with untested corner cases.

In one environmental monitoring system, when asked to enumerate distinct response scenarios, the model focused on high-level alert categories.
In contrast, formal decomposition revealed a combinatorial explosion arising from Boolean constraint boundaries, where different combinations of sensor drift conditions, timing windows, measurement ranges, and pollutant-specific limits produce orders of magnitude more behavioral regions than surface-level structure suggests.

\paragraph{Outcome Precision (0.613).}
This metric exposes substantial imprecision in how LLMs specify constraints.
Mean scores indicate frequent reliance on qualitative descriptions or approximate values rather than exact specifications.
Such imprecision introduces ambiguity into requirements documentation, test case generation, and safety audits.
For example, stating ``temperature must not be too high'' instead of ``temperature \(< 620^\circ\text{C}\)'' prevents consistent implementation, hinders compliance verification, and leaves boundary conditions untested.

\paragraph{Direction Accuracy (0.635).}
Direction Accuracy measures whether LLMs reach correct conclusions, particularly for safety-critical properties where errors have real-world consequences.
This metric uses graduated scoring from \(0.0\) (completely incorrect) to \(1.0\) (correct with sound reasoning), with partial credit awarded for identifying relevant factors despite an incorrect conclusion.

A mean score of \(0.635\) indicates that while LLMs often identify relevant aspects of system behavior, they frequently reach flawed conclusions or provide incomplete reasoning.
The most severe failures occur when LLMs categorically misidentify whether safety properties hold.
In one data protection system, formal verification proved that a critical privacy property always holds, yet the LLM incorrectly concluded that the property could be violated, misinterpreting how boundary conditions propagate through compliance checks.
Similarly, in an autonomous driving system, the LLM claimed that a sensor reliability property could be violated under specific conditions, when verification showed it always holds.

These failures represent fundamental errors about system guarantees—the difference between confidently deploying a system and discovering critical vulnerabilities only after incidents occur.

\paragraph{Coverage Completeness (0.490).}
Coverage Completeness highlights a weak performance area, with LLMs identifying fewer than half of the actual decision scenarios.
This systematic undercounting creates blind spots where critical bugs tend to hide.
The gap arises because LLMs reason at categorical levels while missing how combinations of constraints multiply scenarios.

When asked to enumerate scenarios, models typically identify only the obvious, top-level categories visible in code structure.
However, each category subdivides along Boolean constraint boundaries, where combinations of comparisons, timing windows, and state checks produce distinct behavioral regions.
The resulting combinatorial interaction creates orders of magnitude more scenarios than surface-level reasoning suggests.
This mirrors a common developer pitfall: believing that ``all cases'' have been tested when only the obvious paths have been exercised.

\paragraph{Control Flow Understanding (0.746).}
Control Flow Understanding represents the strongest area of LLM performance.
Models demonstrate relatively strong capability in interpreting branching logic, condition precedence,
and short-circuit behavior.
This is the only metric where mean performance exceeds \(0.7\), suggesting that LLMs can effectively infer common control-flow idioms through code analysis.

Nevertheless, complex override patterns and deeply nested conditionals remain challenging.
Formal methods retain a key advantage by providing deterministic analysis that is independent of surface-level code structure or stylistic variation.

\paragraph{Edge Case Detection (0.597).}
Edge Case Detection shows that LLMs identify fewer than 60\% of edge cases.
Models tend to focus on obvious scenarios while missing counterintuitive behaviors.
In monitoring systems, LLMs typically recognize high-level operational modes but overlook scenarios in which subsystems in non-standard states can still trigger critical actions through indirect pathways,
such as quorum voting.

These counterintuitive interactions—where naive reasoning predicts one outcome but complex
logical dependencies produce another—are precisely where production bugs most often emerge.
Failing to identify such cases leaves test suites weakest exactly where failures are most likely to occur in the field.

\paragraph{Decision Boundary Clarity (0.695).}
Decision Boundary Clarity shows moderate performance. LLMs partially succeed in identifying
constraints that influence system behavior but struggle to extract precise boundary conditions.
Models often recognize that factors such as ``weather adjustments'' or ``sensor drift'' matter,
yet fail to identify the exact constraint expressions where behavior changes—such as specific
comparisons (\(\geq\) vs.\ \(>\)), threshold values (79 vs.\ 80), or logical combinations.

LLMs also miss narrow ``dead zones'' where small ranges of inputs trigger different logic, and
how constraints vary across operational modes. This limitation is critical because implementations
require exact Boolean conditions. Developers need precise constraint expressions to implement,
test boundary behavior, and verify correctness. Region decomposition exhaustively enumerates
these constraints, enabling precise implementation and comprehensive test case generation.

These data are presented in Figures~\ref{fig:mean-by-metric} and Figures~\ref{fig:median-score-chart}.

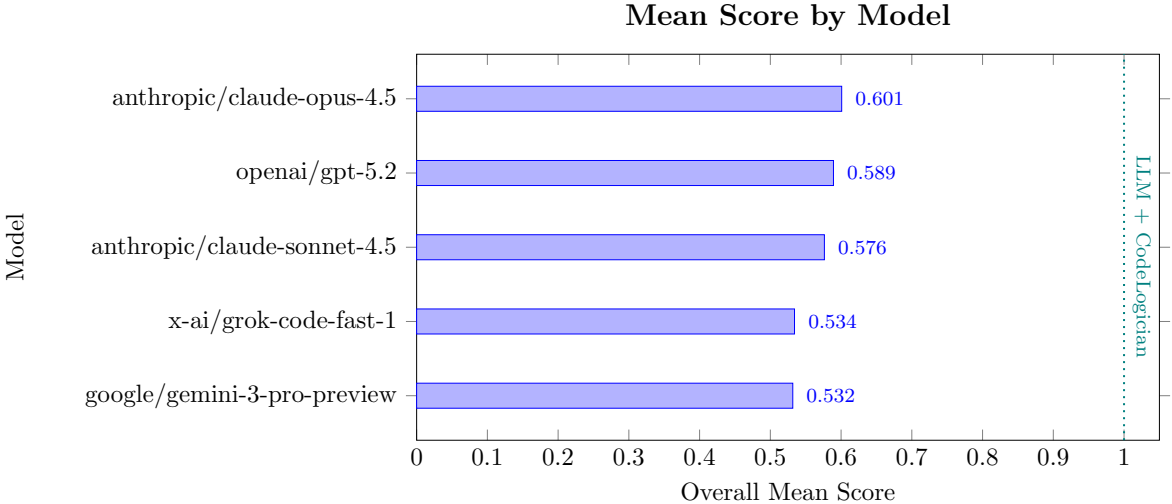
\begin{figure}[htbp]
  \centering
\begin{adjustbox}{max width=\textwidth}
\begin{tikzpicture}

  % Main bar chart
  \begin{axis}[
      title={\large\bfseries Mean Score by Model},
      xbar,
      xmin=0,
      xmax=1.05,
      xlabel={Overall Mean Score},
      ylabel={Model},
      ytick=data,
      yticklabel style={text width=5cm, align=right},
      symbolic y coords={
          google/gemini-3-pro-preview,
          x-ai/grok-code-fast-1,
          anthropic/claude-sonnet-4.5,
          openai/gpt-5.2,
          anthropic/claude-opus-4.5
        },
      enlarge y limits=0.15,
      width=12cm,
      height=7cm,
      nodes near coords,
      nodes near coords align={horizontal},
      every node near coord/.append style={
          font=\footnotesize,
          /pgf/number format/fixed,
          /pgf/number format/precision=3,
          anchor=west,
          xshift=2pt,
        },
    ]
    \addplot table[
        x=overall_average_score,
        y=model,
        col sep=comma,
        row sep=\\,
      ] {
        model,overall_average_score\\
        anthropic/claude-opus-4.5,0.6008632672347839\\
        openai/gpt-5.2,0.5892392243522563\\
        anthropic/claude-sonnet-4.5,0.5764117262396365\\
        x-ai/grok-code-fast-1,0.5341634856947183\\
        google/gemini-3-pro-preview,0.5317321314126643\\
      };
  \end{axis}

  % Overlay axis only for the vertical dotted line and vertical label
  \begin{axis}[
      xmin=0,
      xmax=1.05,
      ymin=0,
      ymax=1.05,
      axis lines=none,
      width=12cm,
      height=7cm,
      ticks=none,
      yticklabels=\empty,
    ]
    \addplot[
      teal,
      thick,
      dotted,
    ] coordinates {(1,0) (1,2)};

    \node[
      teal,
      rotate=270,
      anchor=south,
      font=\footnotesize
    ] at (axis cs:1,0.5) {LLM + CodeLogician};
  \end{axis}

\end{tikzpicture}
\end{adjustbox}
\caption{Mean score by model across seven metrics, relative to the scores when models were augmented with CodeLogician. Claude Opus 4.5 achieves the highest score (0.601), followed by GPT-5.2 (0.589). Despite these differences, all models perform substantially below the level achieved when augmented with formal methods.}
\label{fig:score-by-model}
\end{figure}

\begin{figure}[htbp]
  \centering
\begin{adjustbox}{max width=\textwidth}
\begin{tikzpicture}

  % Main bar chart
  \begin{axis}[
      title={\large\bfseries Mean Score by Metric},
      xbar,
      xmin=0,
      xmax=1.05,
      xlabel={Mean Score},
      ylabel={Metric},
      ytick=data,
      yticklabel style={text width=6cm, align=right},
      symbolic y coords={
          State Space Estimation Accuracy,
          Coverage Completeness,
          Edge Case Detection,
          Outcome Precision,
          Direction Accuracy,
          Decision Boundary Clarity,
          Control Flow Understanding
        },
      enlarge y limits=0.15,
      width=12cm,
      height=10cm,
      nodes near coords,
      nodes near coords align={horizontal},
      every node near coord/.append style={
          font=\footnotesize,
          /pgf/number format/fixed,
          /pgf/number format/precision=3,
          anchor=west,
          xshift=2pt,
        },
    ]
    \addplot table[
        x=average_score,
        y=metric,
        col sep=comma,
        row sep=\\,
      ] {
        metric,average_score\\
        Control Flow Understanding,0.746263888888889\\
        Coverage Completeness,0.48973763662168646\\
        Decision Boundary Clarity,0.6948336041526724\\
        Direction Accuracy,0.6354255555555555\\
        Edge Case Detection,0.5967482746042079\\
        Outcome Precision,0.6129166666666668\\
        State Space Estimation Accuracy,0.18600519079223035\\
      };
  \end{axis}

  % Overlay axis only for the vertical dotted line and vertical label
  \begin{axis}[
      xmin=0,
      xmax=1.05,
      ymin=0,
      ymax=1.05,
      axis lines=none,
      width=12cm,
      height=10cm,
      ticks=none,
      yticklabels=\empty,
    ]
    \addplot[
      teal,
      thick,
      dotted,
    ] coordinates {(1,0) (1,2)};

    \node[
      teal,
      rotate=270,
      anchor=south,
      font=\footnotesize
    ] at (axis cs:1,0.5) {LLM + CodeLogician};
  \end{axis}

\end{tikzpicture}
\end{adjustbox}
\caption{Mean score by metric, relative to performance once augmented with CodeLogician.}
\label{fig:mean-by-metric}
\end{figure}

\begin{figure}[htbp]
  \centering
\begin{adjustbox}{max width=\textwidth}
\begin{tikzpicture}

  % Main bar chart
  \begin{axis}[
      title={\large\bfseries Median Score by Metric},
      xbar,
      xmin=0,
      xmax=1.05,
      xlabel={Median Score},
      ylabel={Metric},
      ytick=data,
      yticklabel style={text width=6cm, align=right},
      symbolic y coords={
          State Space Estimation Accuracy,
          Coverage Completeness,
          Edge Case Detection,
          Outcome Precision,
          Direction Accuracy,
          Decision Boundary Clarity,
          Control Flow Understanding
        },
      enlarge y limits=0.15,
      width=12cm,
      height=10cm,
      nodes near coords,
      nodes near coords align={horizontal},
      every node near coord/.append style={
          font=\footnotesize,
          /pgf/number format/fixed,
          /pgf/number format/precision=3,
          anchor=west,
          xshift=2pt,
        },
    ]
    \addplot table[
        x=median_score,
        y=metric,
        col sep=comma,
        row sep=\\,
      ] {
        metric,median_score\\
        Control Flow Understanding,0.8333333333333334\\
        Coverage Completeness,0.45666666666666667\\
        Decision Boundary Clarity,0.758968253968254\\
        Direction Accuracy,0.7833333333333333\\
        Edge Case Detection,0.5880952380952381\\
        Outcome Precision,0.6645833333333333\\
        State Space Estimation Accuracy,0.0933937105932916\\
      };
  \end{axis}

  % Overlay axis only for the vertical dotted line and vertical label
  \begin{axis}[
      xmin=0,
      xmax=1.05,
      ymin=0,
      ymax=1.05,
      axis lines=none,
      width=12cm,
      height=10cm,
      ticks=none,
      yticklabels=\empty,
    ]
    \addplot[
      teal,
      thick,
      dotted,
    ] coordinates {(1,0) (1,2)};

    \node[
      teal,
      rotate=270,
      anchor=south,
      font=\footnotesize
    ] at (axis cs:1,0.5) {LLM + CodeLogician};
  \end{axis}

\end{tikzpicture}
\end{adjustbox}
\caption{Mean score by metric, relative to performance once augmented with CodeLogician.}
\label{fig:median-score-chart}
\end{figure}

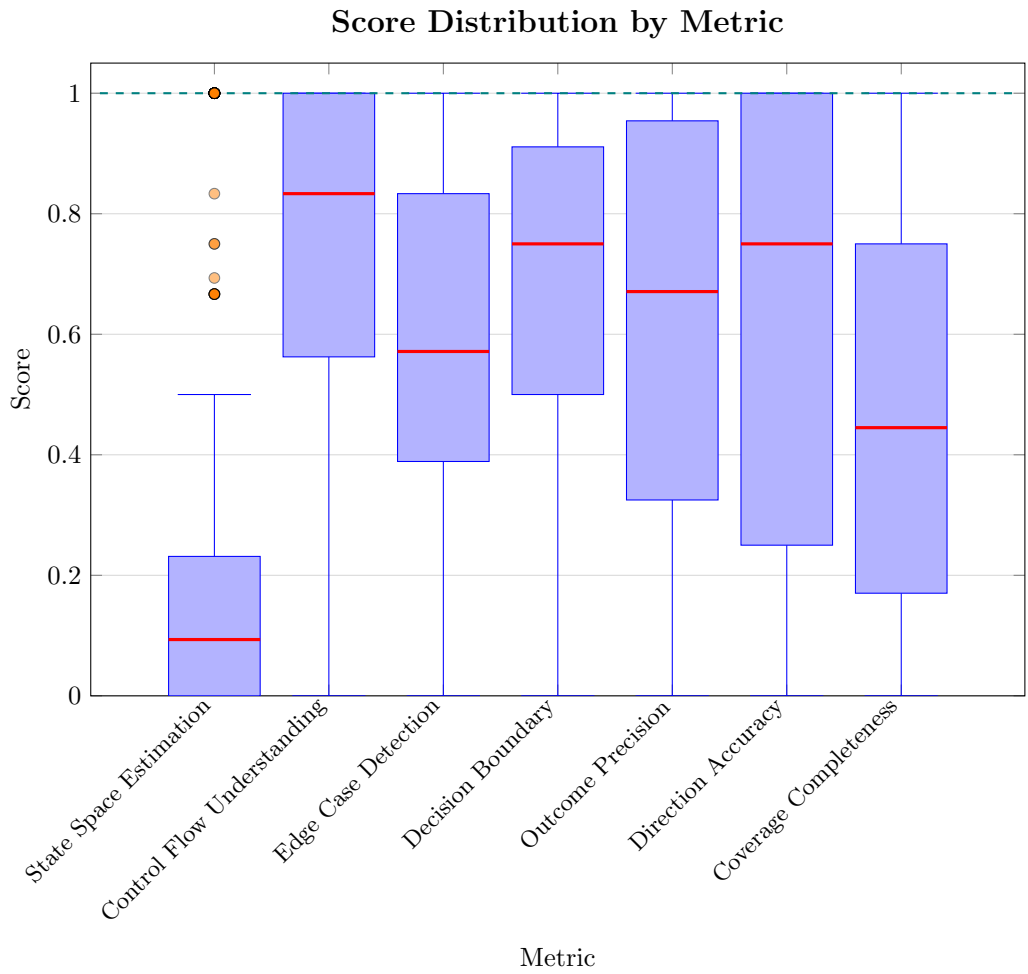
\begin{figure}[htbp]
  \centering
\begin{adjustbox}{max width=\textwidth}
\begin{tikzpicture}
  \begin{axis}[
      title={\large\bfseries Score Distribution by Metric},
      ylabel={Score},
      xlabel={Metric},
      boxplot/draw direction=y,
      cycle list={{draw=blue,fill=blue!30!white, boxplot/every median/.style={red, very thick}}},
      xtick={1,2,3,4,5,6,7},
      xticklabels={
          State Space Estimation,
          Control Flow Understanding,
          Edge Case Detection,
          Decision Boundary,
          Outcome Precision,
          Direction Accuracy,
          Coverage Completeness
        },
      x tick label style={rotate=45, anchor=east, font=\small},
      ymin=0,
      ymax=1.05,
      width=14cm,
      height=10cm,
      ymajorgrids=true,
      grid style={gray!30},
    ]

    \addplot+[boxplot-style] table [y=score] {data/state-space-estimation-accuracy.dat};

    % Control Flow Understanding
    \addplot+[boxplot-style] table [y=score] {data/control-flow-understanding.dat};

    % Edge Case Detection
    \addplot+[boxplot-style] table [y=score] {data/edge-case-detection.dat};

    % Decision Boundary Clarity
    \addplot+[boxplot-style] table [y=score] {data/decision-boundary-clarity.dat};

    % Outcome Precision
    \addplot+[boxplot-style] table [y=score] {data/outcome-precision.dat};

    % Direction Accuracy
    \addplot+[boxplot-style] table [y=score] {data/direction-accuracy.dat};

    % Coverage Completeness
    \addplot+[boxplot-style] table [y=score] {data/coverage-completeness.dat};

    % Horizontal dashed line at y=1.0
    \draw[teal, thick, dashed] (axis cs:0.0,1.0) -- (axis cs:8.0,1.0);

  \end{axis}
\end{tikzpicture}
\end{adjustbox}
\caption{Score distributions across the seven metrics.}
\label{fig:box-plot}
\end{figure}

\subsubsection{Model Comparison by Metric}
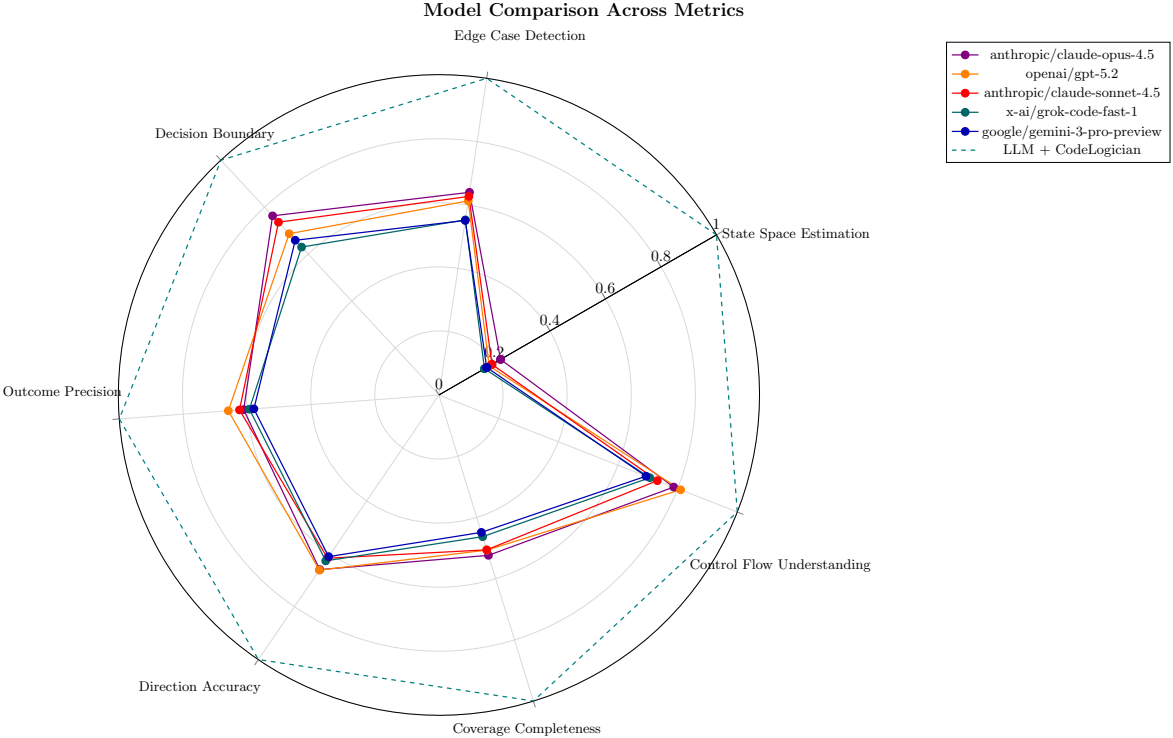
\begin{figure}[htbp]
  \centering
% Radar chart for Model Comparison Across Metrics
\begin{adjustbox}{max width=\textwidth}
\begin{tikzpicture}
  \begin{polaraxis}[
      width=16cm,
      height=16cm,
      rotate=30,
      ytick={0, 0.2, 0.4, 0.6, 0.8, 1.0},
      ymin=0,
      ymax=1.0,
      xtick={0,51.43,102.86,154.29,205.71,257.14,308.57},
      xticklabels={
          State Space Estimation,
          Edge Case Detection,
          Decision Boundary,
          Outcome Precision,
          Direction Accuracy,
          Coverage Completeness,
          Control Flow Understanding
        },
      xticklabel style={font=\small},
      legend style={
          at={(1.3,0.5)},
          anchor=west,
          font=\small
        },
      grid=both,
      major grid style={gray!30},
    ]

    % Reference line at 1.0 (LLM + CodeLogician)
    \addplot[teal, thick, dashed, forget plot] coordinates {
        (0,1.0)
        (51.43,1.0)
        (102.86,1.0)
        (154.29,1.0)
        (205.71,1.0)
        (257.14,1.0)
        (308.57,1.0)
        (360,1.0)
      };

    % Claude Opus 4.5 (purple)
    \addplot[violet, mark=*, thick, mark options={fill=violet, scale=1.2}] coordinates {
        (0,0.222)         % State Space Estimation
        (51.43,0.640)     % Edge Case Detection
        (102.86,0.763)    % Decision Boundary
        (154.29,0.611)    % Outcome Precision
        (205.71,0.660)    % Direction Accuracy
        (257.14,0.524)    % Coverage Completeness
        (308.57,0.787)    % Control Flow Understanding
        (360,0.222)       % Close the polygon
      };
    \addlegendentry{anthropic/claude-opus-4.5}

    % GPT-5.2 (orange)
    \addplot[orange, mark=*, thick, mark options={fill=orange, scale=1.2}] coordinates {
        (0,0.181)         % State Space Estimation
        (51.43,0.613)     % Edge Case Detection
        (102.86,0.687)    % Decision Boundary
        (154.29,0.659)    % Outcome Precision
        (205.71,0.662)    % Direction Accuracy
        (257.14,0.507)    % Coverage Completeness
        (308.57,0.810)    % Control Flow Understanding
        (360,0.181)       % Close the polygon
      };
    \addlegendentry{openai/gpt-5.2}

    % Claude Sonnet 4.5 (red)
    \addplot[red, mark=*, thick, mark options={fill=red, scale=1.2}] coordinates {
        (0,0.191)
        (51.43,0.627)
        (102.86,0.736)
        (154.29,0.624)
        (205.71,0.617)
        (257.14,0.506)
        (308.57,0.732)
        (360,0.191)
      };
    \addlegendentry{anthropic/claude-sonnet-4.5}

    % Grok Code Fast 1 (teal)
    \addplot[teal!80!black, mark=*, thick, mark options={fill=teal!80!black, scale=1.2}] coordinates {
        (0,0.164)
        (51.43,0.553)
        (102.86,0.630)
        (154.29,0.593)
        (205.71,0.627)
        (257.14,0.463)
        (308.57,0.707)
        (360,0.164)
      };
    \addlegendentry{x-ai/grok-code-fast-1}

    % Gemini 3 Pro (blue)
    \addplot[blue!70!black, mark=*, thick, mark options={fill=blue!70!black, scale=1.2}] coordinates {
        (0,0.172)
        (51.43,0.551)
        (102.86,0.659)
        (154.29,0.579)
        (205.71,0.611)
        (257.14,0.449)
        (308.57,0.694)
        (360,0.172)
      };
    \addlegendentry{google/gemini-3-pro-preview}

    % Add legend entry for reference line
    \addlegendimage{teal, thick, dashed}
    \addlegendentry{LLM + CodeLogician}

  \end{polaraxis}

  % Add title at the top (unrotated)
  \node[above] at (current bounding box.north) {\large\bfseries Model Comparison Across Metrics};

\end{tikzpicture}
\end{adjustbox}
\caption{Radar chart showing model performance in the seven metrics tested. Claude Opus 4.5 leads in State Space Estimation and Decision Boundary, while GPT 5.2 outperforms the others on Control Flow Understanding. Despite these variations, all models perform significantly worse than when augmented with CodeLogician.}
\label{fig:radar-chart}
\end{figure}

\begin{figure}[htbp]
  \centering
% Grouped bar chart - Model Comparison Across All Metrics
\begin{adjustbox}{max width=\textwidth}
\begin{tikzpicture}
  \begin{axis}[
      title={\large\bfseries Model Comparison Across All Metrics},
      ybar,
      bar width=0.15cm,
      width=18cm,
      height=12cm,
      ylabel={Score (0-1)},
      xlabel={Metrics},
      ymin=0,
      ymax=1.15,
      symbolic x coords={
          State Space Estimation,
          Control Flow Understanding,
          Edge Case Detection,
          Decision Boundary,
          Outcome Precision,
          Direction Accuracy,
          Coverage Completeness
        },
      xtick=data,
      x tick label style={font=\small, rotate=45, anchor=east},
      legend style={
          at={(0.02,0.98)},
          anchor=north west,
          font=\small,
          draw=black,
          fill=white
        },
      ymajorgrids=true,
      grid style={gray!30},
      enlarge x limits=0.15,
      nodes near coords,
      every node near coord/.append style={
          font=\tiny,
          /pgf/number format/fixed,
          /pgf/number format/precision=2
        },
    ]

    % Claude Opus 4.5 (purple)
    \addplot[fill=violet!70, draw=black] coordinates {
        (State Space Estimation, 0.222)
        (Control Flow Understanding, 0.787)
        (Edge Case Detection, 0.640)
        (Decision Boundary, 0.763)
        (Outcome Precision, 0.611)
        (Direction Accuracy, 0.660)
        (Coverage Completeness, 0.524)
      };
    \addlegendentry{anthropic/claude-opus-4.5}

    % Claude Sonnet 4.5 (red)
    \addplot[fill=red!70, draw=black] coordinates {
        (State Space Estimation, 0.191)
        (Control Flow Understanding, 0.732)
        (Edge Case Detection, 0.627)
        (Decision Boundary, 0.736)
        (Outcome Precision, 0.624)
        (Direction Accuracy, 0.617)
        (Coverage Completeness, 0.506)
      };
    \addlegendentry{anthropic/claude-sonnet-4.5}

    % Gemini 3 Pro (blue)
    \addplot[fill=blue!60, draw=black] coordinates {
        (State Space Estimation, 0.172)
        (Control Flow Understanding, 0.694)
        (Edge Case Detection, 0.551)
        (Decision Boundary, 0.659)
        (Outcome Precision, 0.579)
        (Direction Accuracy, 0.611)
        (Coverage Completeness, 0.449)
      };
    \addlegendentry{google/gemini-3-pro-preview}

    % GPT-5.2 (orange)
    \addplot[fill=orange!80, draw=black] coordinates {
        (State Space Estimation, 0.181)
        (Control Flow Understanding, 0.810)
        (Edge Case Detection, 0.613)
        (Decision Boundary, 0.687)
        (Outcome Precision, 0.659)
        (Direction Accuracy, 0.662)
        (Coverage Completeness, 0.507)
      };
    \addlegendentry{openai/gpt-5.2}

    % Grok Code Fast 1 (teal)
    \addplot[fill=teal!70, draw=black] coordinates {
        (State Space Estimation, 0.164)
        (Control Flow Understanding, 0.707)
        (Edge Case Detection, 0.553)
        (Decision Boundary, 0.630)
        (Outcome Precision, 0.593)
        (Direction Accuracy, 0.627)
        (Coverage Completeness, 0.463)
      };
    \addlegendentry{x-ai/grok-code-fast-1}

    % Horizontal dashed line at y=1.0 extending full width
    \draw[teal, thick, dashed] ({rel axis cs:0,0}|-{axis cs:State Space Estimation,1.0}) --
          ({rel axis cs:1,0}|-{axis cs:State Space Estimation,1.0});

    % Label for the reference line
    \node[teal, font=\small, anchor=west] at (axis cs:Direction Accuracy,1.02)
          {LLM + CodeLogician};

  \end{axis}
\end{tikzpicture}
\end{adjustbox}
\caption{Chart showing model performance in the seven metrics tested. Scores are relative to those of the LLMs when augmented with CodeLogician.}
\label{fig:grouped-bar-chart}
\end{figure}
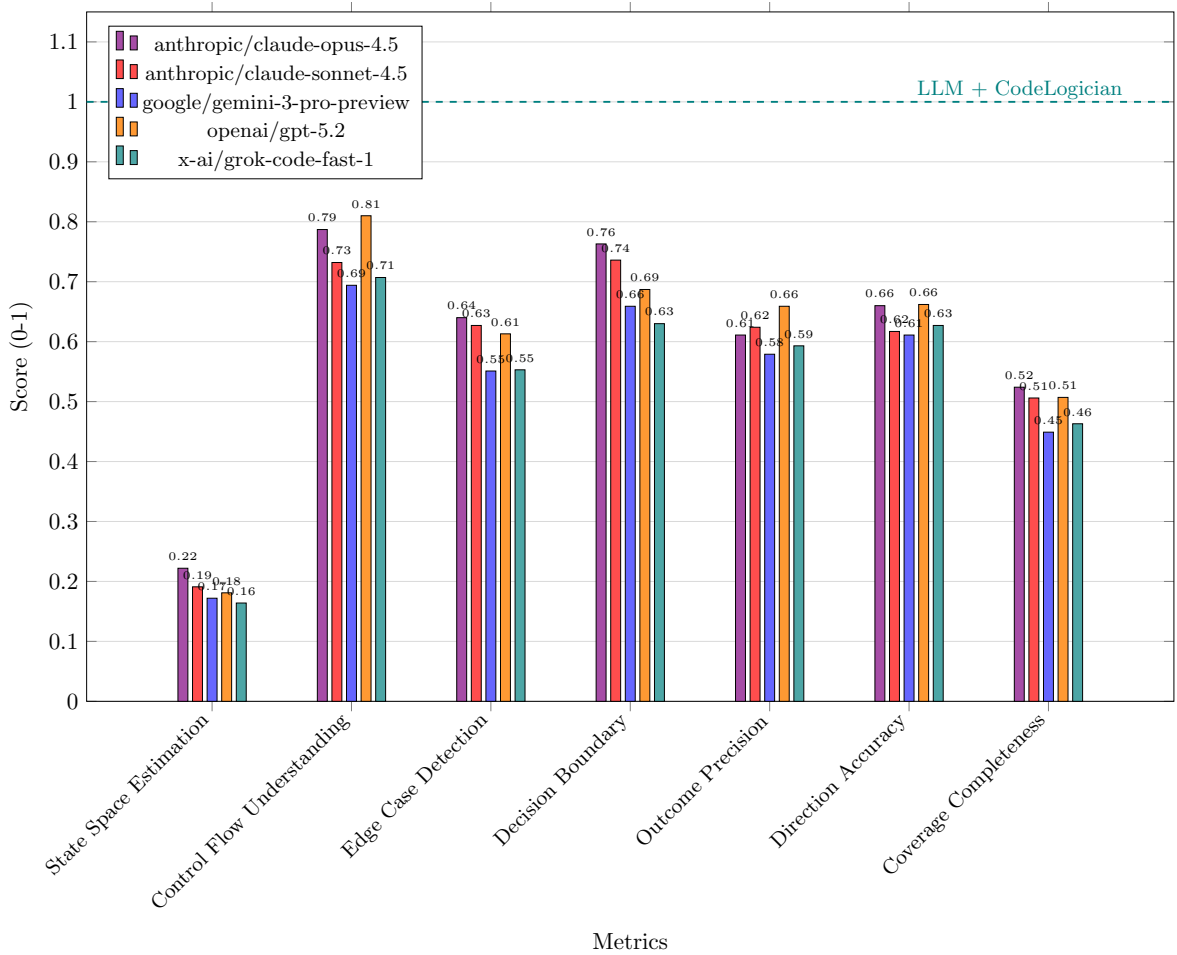

Figures~\ref{fig:radar-chart} and \ref{fig:grouped-bar-chart} show how each model performed on individual metrics:
Claude Opus~4.5 achieves the highest overall performance (\(0.601\)), leading in
\emph{State Space Estimation Accuracy} (\(0.222\)), \emph{Decision Boundary Clarity} (\(0.763\)),
\emph{Edge Case Detection} (\(0.640\)), and \emph{Coverage Completeness} (\(0.524\)).
GPT-5.2 achieves the strongest performance in \emph{Control Flow Understanding} (\(0.810\))
and \emph{Outcome Precision} (\(0.659\)), indicating effective pattern recognition for common
control-flow structures and relatively precise constraint specification.

\paragraph{Shared Limitations Across Models.}
Despite these relative differences, all evaluated models perform similarly on
\emph{State Space Estimation Accuracy}, with scores clustered in the
\(0.164\)--\(0.222\) range. Even Claude Opus~4.5, which achieves the highest score on this metric,
still falls far short of the precision achievable through formal decomposition.
This variance on the most challenging metric indicates that fundamental
architectural limitations affect current LLMs in a broadly similar manner when operating
without formal reasoning tools.

\paragraph{Impact of Formal Methods Augmentation.}
The overall performance gap of \(40\)--\(47\) percentage points—comparing LLM-only performance
(\(0.53\)--\(0.60\)) against perfect scores (\(1.0\)) achieved with region decomposition and
verification—quantifies the value added by formal methods tools such as ImandraX when
augmenting LLM-based code analysis.

\paragraph{Reproducibility.}
All raw data and scripts to reproduce these results are available at \url{https://github.com/imandra-ai/code-logic-bench}.

\section{Case Studies}
\subsection{LSE GTT Order Expiry}
% Good Till Time (GTT)
This example concerns itself with an extended (by Imandra) version of the London Stock Exchange trading system specification. The Guide to the Trading System (MIT201) \cite{MIT201} specifies:
\begin{displayquote}
``Any GTT [Good Till Time] orders with an expiry time during any auction call phase will not be expired until after uncrossing has completed and are therefore eligible to participate in that uncrossing.''
\end{displayquote}
The model implements a state machine with two trading modes (Continuous and Auction Call), GTT order lifecycle management, and an auction protection mechanism.
The auction protection logic is implemented in the following Python function, with the full program available in Appendix~\ref{appendix:lse-gtt}.
\begin{lstlisting}[language=python]
def maybe_expire_gtt_order_v2(state: State) -> State:
    """
    Handle GTT order expiry with auction protection.
    """
    if state.gtt_order is None:
        return state
 
    expires_at = state.gtt_order.expires_at
 
    if isinstance(state.mode, AuctionCallMode):
        expires_at = max(expires_at, state.mode.uncross_at + 1)
 
    if state.time >= expires_at:
        return expire_order(state)
    else:
        return state
\end{lstlisting}

\subsubsection{Verification}
The critical property to verify is that auction uncross and order expiry events can never occur simultaneously.
CodeLogician expressed this as a conflict detection predicate in IML:

\begin{lstlisting}[language=ocaml]
let conflict_reachable msgs state k =
  let final_state = run state (List.take k msgs) in
  final_state.auction_event = Uncrossed && final_state.order_event = Expired
\end{lstlisting}
The verification goal asserts that for all valid initial states and message sequences, \lstinline|conflict\_reachable| should return \lstinline|false|.
CodeLogician refuted the verification goal, producing a concrete counterexample:
\begin{lstlisting}[language=ocaml]
let msgs = [Tick; Tick]
let state = {
  time = 2699;
  mode = AuctionCallMode {uncross_at = 2700};
  gtt_order = Some {expires_at = 2700};
  market_order_present_after_uncross = true;
  market_order_extension_period = 1
}
\end{lstlisting}
This sequence of events is illustrated in Figure~\ref{fig:sequence-diagram} (Appendix~\ref{appendix:lse-gtt}).

The counterexample reveals a subtle interaction between two features: the auction protection logic extends order expiry to \lstinline|uncross\_at + 1|, while the market order extension feature delays the actual uncross to \lstinline|uncross\_at + extension\_period|.
When \lstinline|extension\_period = 1|, both events occur at the same tick.

This case demonstrates formal verification's ability to catch feature interaction bugs: the auction protection and market order extension features each work correctly in isolation, but their interaction was not properly considered.
Traditional testing would likely miss this edge case, as it requires specific timing conditions that are difficult to anticipate manually.

\subsection{London Stock Exchange Fee Schedule}
This case study applies CodeLogician to the London Stock Exchange Trading Services Price List, verifying fee calculation logic with mathematical certainty.
For algorithmic trading firms, accurate fee prediction is essential: a fraction of a basis point can determine whether a strategy is profitable, and errors across millions of daily orders compound into significant revenue leakage or unexpected costs.
Exchange fee schedules present challenges: tiered pricing, minimum floors, rebates, and surcharges can interact in complex ways, and specifications are published as prose documents rather than machine-readable formats.
Traditional testing struggles to cover all combinations exhaustively.

\subsubsection{Verified Invariants}
CodeLogician formally verified two invariants against all possible inputs.

The first invariant, ``Non-Persistent Order Charge Isolation,'' establishes that an Immediate or Cancel (IOC) or Fill or Kill (FOK) order costs exactly £0.01 more than an identical Day (DAY) order, due to the order entry charge.
CodeLogician expressed this property in IML as:

\begin{lstlisting}[language=ocaml]
fun (price : real) (quantity : int) (is_aggressive : bool) (is_hidden : bool) 
    (package : package_type) (lps_tier : lps_tier_type) (security_type : security_type) 
    (cumulative_value : real) ->
  calculate_total_cost price quantity true is_aggressive is_hidden package lps_tier 
    security_type cumulative_value 
  = calculate_total_cost price quantity false is_aggressive is_hidden package lps_tier 
      security_type cumulative_value +. 0.01
\end{lstlisting}
CodeLogician proved this property holds for all combinations of inputs, demonstrating that the order entry charge logic is completely isolated from all other fee components.

The second invariant, ``LPS Zero-Fee Execution,'' guarantees that when the rate equals 0.00bp and the minimum charge equals £0.00, the execution fee is exactly zero:

\begin{lstlisting}[language=ocaml]
fun (notional_value : real) -> 
  calculate_execution_fee notional_value 0.0 0.0 false = 0.0
\end{lstlisting}

This ensures that liquidity providers---who receive a 0.00bp rate with no minimum---are never accidentally charged due to floating-point errors or logic edge cases.

\subsubsection{Region Decomposition}
Region decomposition revealed 6 distinct behavioral regions in the execution fee function, arising from the combinatorial interaction of factors including hidden order premiums, minimum charge floors, and negative rebate rates.
The regions are partitioned by whether the order is hidden (triggering the +0.25bp premium), whether the rate is non-negative, and whether the minimum charge floor dominates the calculated fee.

The analysis identified ``crossover notionals''---trade sizes below which the minimum charge dominates the calculated fee.
For example, a £1,000 trade at the standard 0.45bp rate yields a calculated fee of £0.045, but the minimum floor of £0.11 applies, resulting in an effective rate of 11bp---24 times the stated rate.
Such boundaries vary by pricing scheme: the crossover for Standard Tier 1 (0.45bp, 11p minimum) is £2,444.44, while Package 1 (0.15bp, 5p minimum) crosses over at £3,333.33.

Negative rates (LPS rebates) bypass the minimum charge logic entirely---a design decision that prevents floors from being applied to rebates.
CodeLogician verified an edge case where a large hidden order with LPS FTSE100 Top Tier (-0.15bp rebate + 0.25bp hidden premium) pays a net 0.10bp, not zero.

While this particular example may seem straightforward, the same analysis, when applied to complex fee schedules, can reveal hidden complexity and compliance risks, and provide a mathematical foundation for trust in the system.

\subsection{Multilateral Netting Engine}
In this example, CodeLogician analyzes a simple multilateral netting engine to find two critical bugs.
The system has two critical invariants:
\begin{enumerate}
    \item Zero-sum Conservation – the sum of all net positions must equal zero
    \item Netting Efficiency – no party's exposure must increase as a result of netting.
\end{enumerate}
Formal methods found issues pertaining to both of these requirements in the code.
Bugs were found before any testing, concrete counterexamples presented, and once the code was corrected, CodeLogician proved the code mathematically correct.
An excerpt of the code is exhibited below, with the full program available in Appendix~\ref{appendix:mne-gtt}.

\begin{lstlisting}[language=python]
def calculate_net_obligations(trades: List[Trade]) -> Dict[str, float]:
    """
    Calculate net obligations for all participants in a multilateral netting cycle.
    """
    net_positions: Dict[str, float] = {}
    
    for trade in trades:
        net_positions[trade.payer_id] = \
            net_positions.get(trade.payer_id, 0.0) - trade.amount
        
        net_positions[trade.receiver_id] = \
            net_positions.get(trade.receiver_id, 0.0) + trade.amount
    
    return net_positions
\end{lstlisting}

\subsubsection{Input Validation}
The first bug to be found was an input validation problem where negative values were accepted, causing a violation of the netting efficiency requirement.
CodeLogician found the counterexample

\begin{lstlisting}[language=ocaml]
Trade(payer_id="", receiver_id="", amount=-10834.0)
\end{lstlisting}

In this case, Net Exposure, $\lvert10834\rvert + \lvert-10834\rvert = 21668$, is greater than Gross Exposure, $\lvert-10834\rvert = 10834$.

\subsubsection{Floating Point Precision}
The second issue found was with the use of floating-point arithmetic.
CodeLogician identified that floating-point errors would cause the zero-sum conservation requirement to be violated.
IEEE 754 \cite{IEEE754} binary representation cannot exactly represent most decimal values; accumulated errors compound with each operation.

Consider the classic demonstration: summing 0.1 ten times in floating-point yields 0.9999999999999999, not 1.0.
In high-frequency clearing with thousands of trades per second, such errors accumulate to material discrepancies.
CodeLogician's counterexample demonstrated a scenario where\\
\lstinline|{sum(net_positions.values()) != 0.0|, violating the zero-sum invariant despite mathematically balanced trades.

The fix required migrating from Python's \lstinline|float| to \lstinline|Decimal| \cite{Python-Decimal} for arbitrary-precision arithmetic, combined with defense-in-depth validation within the core algorithm.
After applying the fix, CodeLogician re-verified the implementation, proving all three verification goals: exact zero-sum conservation, tolerance-based zero-sum, and netting efficiency.
The corrected implementation provides mathematical certainty of correctness---a requirement for systems subject to EMIR \cite{EMIR} and Dodd-Frank Title VII \cite{DoddFrank}.

\section{Conclusion and Future Work}
\label{sec:conclusion}

We have introduced \emph{CodeLogician}, a neuro-symbolic agent and framework for rigorous reasoning about software systems that integrates Large Language Models (LLMs) with automated reasoning engines.
We argue that while LLMs excel at translation, abstraction, and interaction, they fundamentally lack the capacity for precise, exhaustive reasoning about program behavior.
\emph{CodeLogician} addresses this limitation by teaching LLM-driven agents to construct explicit formal models of software and by delegating semantic reasoning to formal tools such as \emph{ImandraX}.

A central contribution of this work is the demonstration that formal augmentation fundamentally changes the reasoning capabilities of LLM-based systems.
Using a newly introduced benchmark targeting mathematical reasoning about software logic, we showed that LLM-only approaches systematically fail on tasks requiring exact boundary identification, combinatorial reasoning, and exhaustive coverage.
By contrast, LLMs augmented with \emph{CodeLogician} achieve complete and precise analysis, closing a 41--47 percentage point gap across multiple rigorously defined metrics.
These results provide empirical evidence that rigorous program analysis cannot be achieved by statistical reasoning alone and that formal reasoning engines are essential for scaling AI-assisted software development beyond heuristic methods.

Beyond empirical results, we presented \emph{CodeLogician} as a general framework rather than a single agent or solver.
Its reasoner-agnostic architecture, agentic orchestration layer, and support for continuous, project-scale autoformalization enable principled integration of multiple reasoning tools and workflows.
The \emph{CodeLogician} server and \emph{PyIML} strategies demonstrate how formalization can be performed incrementally and persistently over evolving codebases, making formal reasoning practical in real-world development environments.

\paragraph{Future Work.}
Applying LLMs to formal reasoning about software presents both substantial challenges and exceptional opportunities.
LLMs remain imperfect at maintaining global consistency, respecting strict semantic constraints, and scaling reasoning across large, interdependent codebases.
At the same time, their strengths in translation, abstraction, and interaction make them uniquely well suited to act as interfaces between informal software artifacts and formal mathematical models.

Our ongoing work in this direction is embodied in \emph{SpecLogician}, a complementary project focused on scalable, data-driven formal specification and refinement for large software systems.
\emph{SpecLogician} extends the ideas introduced in this paper by shifting the emphasis from individual code fragments to collections of specifications, traces, tests, logs, and other real-world artifacts.
By combining LLMs with automated reasoning, \emph{SpecLogician} aims to incrementally synthesize, refine, and validate formal specifications at scale, even in settings where complete or authoritative specifications do not exist upfront.

Looking forward, we plan to deepen the integration between \emph{CodeLogician} and \emph{SpecLogician}, enabling closed-loop workflows in which specifications, implementations, and empirical artifacts co-evolve under formal constraints.
Additional future directions include extending support to new source languages and reasoning backends, improving dependency-aware and incremental formalization strategies, and using formal counterexamples, region decompositions, and proofs to actively guide LLM behavior during development.

Ultimately, we view \emph{CodeLogician} and \emph{SpecLogician} as steps toward a new class of AI-assisted software engineering systems in which LLMs and formal reasoning engines operate symbiotically.
By grounding AI-assisted development in explicit formal models and mathematically sound analysis, such systems can move beyond probabilistic correctness toward rigorous, auditable, and scalable reasoning about complex software systems.

\newpage
\appendix
\section{LSE GTT Order Expiry}\label{appendix:lse-gtt}

\subsection{Python model}\
\begin{lstlisting}[language=python]
"""
LSE GTT (Good Till Time) Order Expiry Model

This model captures the interaction between Good Till Time orders and 
auction uncross events, as described in the London Stock Exchange Guide 
to the Trading System (MIT201).

Key business rule from MIT201:
"Any GTT orders with an expiry time during any auction call phase will 
not be expired until after uncrossing has completed and are therefore 
eligible to participate in that uncrossing."
"""

from dataclasses import dataclass
from typing import Optional, List
from enum import Enum


# Type definitions
Time = int  # Nanoseconds since midnight


class AuctionEvent(Enum):
    """Events emitted when the model changes mode."""
    AUCTION_CALL_STARTED = "auction_call_started"
    UNCROSSED = "uncrossed"


class OrderEvent(Enum):
    """Events emitted when GTT order state changes."""
    CREATED = "created"
    EXPIRED = "expired"


@dataclass
class ContinuousMode:
    """Venue is in continuous trading mode with a scheduled auction."""
    start_auction_call_at: Time


@dataclass
class AuctionCallMode:
    """Venue is in auction call mode with a scheduled uncross."""
    uncross_at: Time


# Union type for mode
Mode = ContinuousMode | AuctionCallMode


@dataclass
class GTTOrder:
    """A Good Till Time order with an expiry time."""
    expires_at: Time


@dataclass
class State:
    """The complete state of the trading venue."""
    time: Time
    auction_call_duration: Time
    auction_interval: Time
    mode: Mode
    gtt_order: Optional[GTTOrder]
    auction_event: Optional[AuctionEvent]
    order_event: Optional[OrderEvent]
    # Extended state for market order feature
    market_order_present_after_uncross: bool
    market_order_extension_period: Time


class Message(Enum):
    """Messages that can be processed by the venue."""
    TICK = "tick"
    

@dataclass
class CreateGTTOrder:
    """Message to create a GTT order."""
    expires_at: Time


# Helper functions

def is_valid_state(state: State) -> bool:
    """
    Check if a state is valid according to business invariants.
    
    VG: State validity invariants should always hold:
    - Time is non-negative
    - Durations are positive
    - Scheduled times are in the future
    - GTT order expiry is in the future
    """
    if state.time < 0:
        return False
    if state.auction_call_duration <= 0:
        return False
    if state.auction_interval <= 0:
        return False
    if state.market_order_extension_period <= 0:
        return False
    
    # Check mode-specific invariants
    if isinstance(state.mode, ContinuousMode):
        if state.mode.start_auction_call_at <= state.time:
            return False
    elif isinstance(state.mode, AuctionCallMode):
        if state.mode.uncross_at <= state.time:
            return False
    
    # GTT order expiry must be in the future
    if state.gtt_order is not None:
        if state.gtt_order.expires_at <= state.time:
            return False
    
    return True


def events_none(state: State) -> bool:
    """Check if no events are set (used for initial states)."""
    return state.auction_event is None and state.order_event is None


# Core business logic functions

def maybe_create_gtt_order(expires_at: Time, state: State) -> State:
    """
    Handle an incoming GTT order.
    
    VG: Orders with expiry <= current time should be rejected
    """
    if expires_at <= state.time:
        # Reject order - no state change
        return state
    else:
        # Accept order
        return State(
            time=state.time,
            auction_call_duration=state.auction_call_duration,
            auction_interval=state.auction_interval,
            mode=state.mode,
            gtt_order=GTTOrder(expires_at=expires_at),
            auction_event=state.auction_event,
            order_event=OrderEvent.CREATED,
            market_order_present_after_uncross=state.market_order_present_after_uncross,
            market_order_extension_period=state.market_order_extension_period
        )


def start_auction(state: State) -> State:
    """
    Start an auction call phase.
    
    Sets mode to AuctionCall, schedules uncross, emits event.
    """
    return State(
        time=state.time,
        auction_call_duration=state.auction_call_duration,
        auction_interval=state.auction_interval,
        mode=AuctionCallMode(uncross_at=state.time + state.auction_call_duration),
        gtt_order=state.gtt_order,
        auction_event=AuctionEvent.AUCTION_CALL_STARTED,
        order_event=state.order_event,
        market_order_present_after_uncross=state.market_order_present_after_uncross,
        market_order_extension_period=state.market_order_extension_period
    )


def uncross(state: State) -> State:
    """
    Complete the auction uncross.
    
    Sets mode to Continuous, schedules next auction, emits event.
    """
    return State(
        time=state.time,
        auction_call_duration=state.auction_call_duration,
        auction_interval=state.auction_interval,
        mode=ContinuousMode(start_auction_call_at=state.time + state.auction_interval),
        gtt_order=state.gtt_order,
        auction_event=AuctionEvent.UNCROSSED,
        order_event=state.order_event,
        market_order_present_after_uncross=state.market_order_present_after_uncross,
        market_order_extension_period=state.market_order_extension_period
    )


def maybe_uncross(uncross_at: Time, state: State) -> State:
    """
    Handle uncrossing with market order extension logic.
    
    VG: If market orders remain, auction is extended by extension_period
    """
    if state.market_order_present_after_uncross:
        # Market orders remain - check if extension period has elapsed
        if state.time >= uncross_at + state.market_order_extension_period:
            # Extension period complete - uncross and clear market order flag
            new_state = State(
                time=state.time,
                auction_call_duration=state.auction_call_duration,
                auction_interval=state.auction_interval,
                mode=state.mode,
                gtt_order=state.gtt_order,
                auction_event=state.auction_event,
                order_event=state.order_event,
                market_order_present_after_uncross=False,
                market_order_extension_period=state.market_order_extension_period
            )
            return uncross(new_state)
        else:
            # Still in extension period
            return state
    else:
        # No market orders - uncross immediately
        return uncross(state)


def maybe_change_mode(state: State) -> State:
    """
    Handle timed mode changes (auction start or uncross).
    
    VG: Mode changes occur at scheduled times
    """
    if isinstance(state.mode, ContinuousMode):
        if state.time >= state.mode.start_auction_call_at:
            return start_auction(state)
    elif isinstance(state.mode, AuctionCallMode):
        if state.time >= state.mode.uncross_at:
            return maybe_uncross(state.mode.uncross_at, state)
    
    return state


def maybe_expire_gtt_order_v2(state: State) -> State:
    """
    Handle GTT order expiry with auction protection.
    
    CRITICAL FIX: Per MIT201, GTT orders expiring during auction call
    should not expire until AFTER uncross completes.
    
    VG: GTT orders with expiry during auction call must not expire
        until after uncross (expires_at is extended to uncross_at + 1)
    """
    if state.gtt_order is None:
        return state
    
    expires_at = state.gtt_order.expires_at
    
    # Key fix: Delay expiry if we're in auction call mode
    if isinstance(state.mode, AuctionCallMode):
        # Extend expiry to after uncross
        expires_at = max(expires_at, state.mode.uncross_at + 1)
    
    if state.time >= expires_at:
        # Expire the order
        return State(
            time=state.time,
            auction_call_duration=state.auction_call_duration,
            auction_interval=state.auction_interval,
            mode=state.mode,
            gtt_order=None,
            auction_event=state.auction_event,
            order_event=OrderEvent.EXPIRED,
            market_order_present_after_uncross=state.market_order_present_after_uncross,
            market_order_extension_period=state.market_order_extension_period
        )
    else:
        return state


def tick(state: State) -> State:
    """
    Process a clock tick - advance time and handle timed events.
    
    Order of operations matters:
    1. Advance time
    2. Check for order expiry
    3. Check for mode changes
    
    VG: Time advances monotonically
    """
    # Advance time
    state = State(
        time=state.time + 1,
        auction_call_duration=state.auction_call_duration,
        auction_interval=state.auction_interval,
        mode=state.mode,
        gtt_order=state.gtt_order,
        auction_event=state.auction_event,
        order_event=state.order_event,
        market_order_present_after_uncross=state.market_order_present_after_uncross,
        market_order_extension_period=state.market_order_extension_period
    )
    
    # Check for order expiry
    state = maybe_expire_gtt_order_v2(state)
    
    # Check for mode changes
    state = maybe_change_mode(state)
    
    return state


def step(msg: Message | CreateGTTOrder, state: State) -> State:
    """
    Process a single message and return the new state.
    
    VG: Each message type should transition state correctly
    """
    if msg == Message.TICK:
        return tick(state)
    elif isinstance(msg, CreateGTTOrder):
        return maybe_create_gtt_order(msg.expires_at, state)
    else:
        return state


def run(state: State, msgs: List[Message | CreateGTTOrder]) -> State:
    """
    Process a sequence of messages.
    
    VG: Event fields are cleared before processing each message
    VG: Message sequence processing is deterministic
    """
    for msg in msgs:
        # Clear events before processing each message
        state = State(
            time=state.time,
            auction_call_duration=state.auction_call_duration,
            auction_interval=state.auction_interval,
            mode=state.mode,
            gtt_order=state.gtt_order,
            auction_event=None,
            order_event=None,
            market_order_present_after_uncross=state.market_order_present_after_uncross,
            market_order_extension_period=state.market_order_extension_period
        )
        state = step(msg, state)
    
    return state


def conflict_reachable(msgs: List[Message | CreateGTTOrder], state: State, k: int = 5) -> bool:
    """
    CRITICAL VERIFICATION GOAL:
    
    Check if it's possible to reach a state where BOTH:
    - An uncross event occurs AND
    - A GTT order expires
    
    at the same time.
    
    Per MIT201, this should NEVER happen - GTT orders expiring during
    auction call must wait until after uncross.
    
    VG: verify that NOT conflict_reachable for all valid initial states
        and message sequences
    """
    # Limit message sequence length
    limited_msgs = msgs[:k]
    
    # Run the model
    final_state = run(state, limited_msgs)
    
    # Check for the conflict
    return (final_state.auction_event == AuctionEvent.UNCROSSED and 
            final_state.order_event == OrderEvent.EXPIRED)


# Verification Goals (VGs) explained in comments above each function
# These are the properties the author was verifying:

# VG1: State validity invariants (is_valid_state)
# VG2: Order rejection logic (maybe_create_gtt_order) 
# VG3: Auction timing correctness (start_auction, uncross)
# VG4: Market order extension logic (maybe_uncross)
# VG5: Mode transition correctness (maybe_change_mode)
# VG6: GTT order expiry protection during auction (maybe_expire_gtt_order_v2)
# VG7: Time monotonicity (tick)
# VG8: Message processing determinism (step, run)
# VG9: CRITICAL - No simultaneous uncross and expiry (conflict_reachable)


\end{lstlisting}
\newpage
\subsection{Sequence diagram}
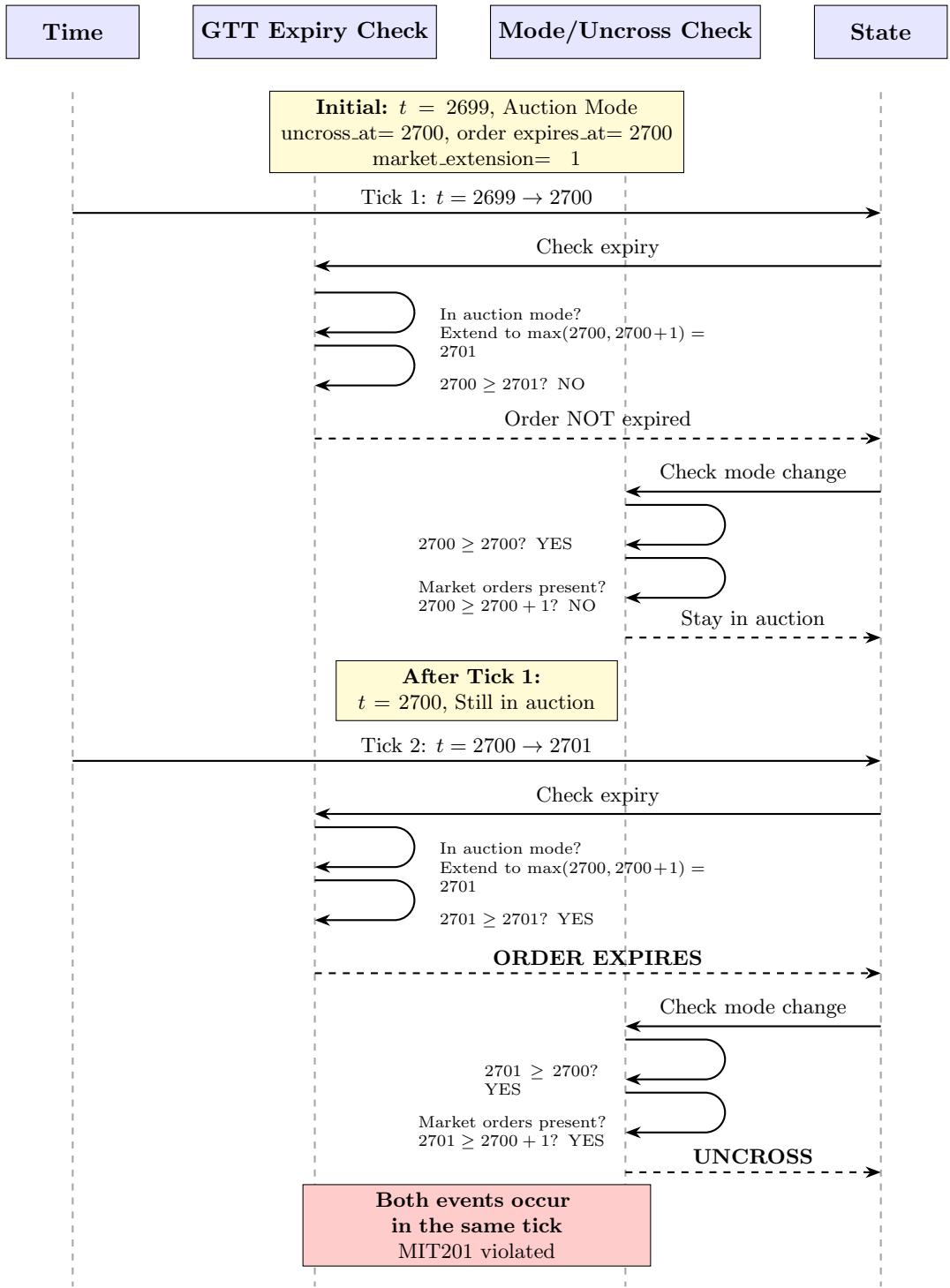
\begin{figure}[h!]
    \centering
    \begin{tikzpicture}[
        node distance=0.8cm,
        participant/.style={
            rectangle,
            minimum width=2cm,
            minimum height=0.8cm,
            text centered,
            draw=black,
            fill=blue!10,
            font=\bfseries
        },
        lifeline/.style={
            thick,
            dashed,
            draw=gray!70
        },
        message/.style={
            ->,
            thick,
            >=Stealth
        },
        selfmsg/.style={
            ->,
            thick,
            >=Stealth,
            rounded corners=3mm
        },
        note/.style={
            rectangle,
            fill=yellow!20,
            draw=black,
            text width=6cm,
            align=left,
            font=\small
        }
    ]
    
    % Participants
    \node[participant] (T) {Time};
    \node[participant, right=of T] (E) {GTT Expiry Check};
    \node[participant, right=of E] (M) {Mode/Uncross Check};
    \node[participant, right=of M] (S) {State};
    
    % % Labels below participants
    % \node[below=0.1cm of T, font=\small] {Time};
    % \node[below=0.1cm of E, font=\small, text width=3cm, align=center] {GTT Expiry Check};
    % \node[below=0.1cm of M, font=\small, text width=2.5cm, align=center] {Mode/Uncross Check};
    % \node[below=0.1cm of S, font=\small] {State};
    
    % Lifelines
    \coordinate (T-start) at ($(T.south) + (0, -0.5)$);
    \coordinate (E-start) at ($(E.south) + (0, -0.5)$);
    \coordinate (M-start) at ($(M.south) + (0, -0.5)$);
    \coordinate (S-start) at ($(S.south) + (0, -0.5)$);
    
    \draw[lifeline] (T-start) -- ++(0, -18) coordinate (T-end);
    \draw[lifeline] (E-start) -- ++(0, -18) coordinate (E-end);
    \draw[lifeline] (M-start) -- ++(0, -18) coordinate (M-end);
    \draw[lifeline] (S-start) -- ++(0, -18) coordinate (S-end);
    
    % Initial y position
    \coordinate (y) at (T-start);
    
    % Initial note
    \coordinate (y) at ($(y) + (0, -1.2)$);
    \node[note, align=center] at ($(T-start)!0.5!(S-start) + (0, -0.6)$) (note1) {
        \textbf{Initial:} $t=2699$, Auction Mode\\
        uncross\_at$=2700$, order expires\_at$=2700$\\
        market\_extension$=1$
    };
    
    % Tick 1
    \coordinate (y) at ($(note1.south) + (0, -0.6)$);
    \draw[message] (T-start |- y) -- node[above, font=\small] {Tick 1: \(t=2699 \to 2700\)} (S-start |- y);
    
    \coordinate (y) at ($(y) + (0, -0.8)$);
    \draw[message] (S-start |- y) -- node[above, font=\small] {Check expiry} (E-start |- y);
    
    % Self messages for E
    \coordinate (y) at ($(y) + (0, -0.4)$);
    \draw[selfmsg] (E-start |- y) -- ++(1.5, 0) -- ++(0, -0.6) -- node[right, font=\scriptsize, text width=4cm, align=left, xshift=1cm] {In auction mode?\\Extend to \(\max(2700, 2700+1)=2701\)} ++(-1.5, 0);
    
    \coordinate (y) at ($(y) + (0, -0.8)$);
    \draw[selfmsg] (E-start |- y) -- ++(1.5, 0) -- ++(0, -0.6) -- node[right, font=\scriptsize, text width=3cm, xshift=1cm] {\(2700 \geq 2701\)? NO} ++(-1.5, 0);
    
    \coordinate (y) at ($(y) + (0, -1.4)$);
    \draw[message, dashed] (E-start |- y) -- node[above, font=\small] {Order NOT expired} (S-start |- y);
    
    % Mode check
    \coordinate (y) at ($(y) + (0, -0.8)$);
    \draw[message] (S-start |- y) -- node[above, font=\small] {Check mode change} (M-start |- y);
    
    \coordinate (y) at ($(y) + (0, -0.2)$);
    \draw[selfmsg] (M-start |- y) -- ++(1.5, 0) -- ++(0, -0.6) -- node[right, font=\scriptsize, text width=3cm, xshift=-4cm] {\(2700 \geq 2700\)? YES} ++(-1.5, 0);
    
    \coordinate (y) at ($(y) + (0, -0.8)$);
    \draw[selfmsg] (M-start |- y) -- ++(1.5, 0) -- ++(0, -0.6) -- node[right, font=\scriptsize, text width=3cm, align=left, xshift=-4cm] {Market orders present?\\\(2700 \geq 2700+1\)? NO} ++(-1.5, 0);
    
    \coordinate (y) at ($(y) + (0, -1.2)$);
    \draw[message, dashed] (M-start |- y) -- node[above, font=\small] {Stay in auction} (S-start |- y);
    
    % Note after Tick 1
    \coordinate (y) at ($(y) + (0, -0.8)$);
    \node[note, text width=4cm, align=center] at ($(T-start |- y)!0.5!(S-start |- y)$) (note2) {
        \textbf{After Tick 1:} \break \(t=2700\), Still in auction
    };
    
    % Tick 2
    \coordinate (y) at ($(note2.south) + (0, -0.6)$);
    \draw[message] (T-start |- y) -- node[above, font=\small] {Tick 2: \(t=2700 \to 2701\)} (S-start |- y);
    
    \coordinate (y) at ($(y) + (0, -0.8)$);
    \draw[message] (S-start |- y) -- node[above, font=\small] {Check expiry} (E-start |- y);
    
    % Self messages for E (Tick 2)
    \coordinate (y) at ($(y) + (0, -0.2)$);
    \draw[selfmsg] (E-start |- y) -- ++(1.5, 0) -- ++(0, -0.6) -- node[right, font=\scriptsize, text width=4cm, align=left, xshift=1cm] {In auction mode?\\Extend to \(\max(2700, 2700+1)=2701\)} ++(-1.5, 0);
    
    \coordinate (y) at ($(y) + (0, -0.8)$);
    \draw[selfmsg] (E-start |- y) -- ++(1.5, 0) -- ++(0, -0.6) -- node[right, font=\scriptsize, text width=3cm, xshift=1cm] {\(2701 \geq 2701\)? YES} ++(-1.5, 0);
    
    \coordinate (y) at ($(y) + (0, -1.4)$);
    \draw[message, dashed] (E-start |- y) -- node[above, font=\small] {\textbf{ORDER EXPIRES}} (S-start |- y);
    
    % Mode check (Tick 2)
    \coordinate (y) at ($(y) + (0, -0.8)$);
    \draw[message] (S-start |- y) -- node[above, font=\small] {Check mode change} (M-start |- y);
    
    \coordinate (y) at ($(y) + (0, -0.2)$);
    \draw[selfmsg] (M-start |- y) -- ++(1.5, 0) -- ++(0, -0.6) -- node[right, font=\scriptsize, text width=2cm, xshift=-3cm] {\(2701 \geq 2700\)? YES} ++(-1.5, 0);
    
    \coordinate (y) at ($(y) + (0, -0.8)$);
    \draw[selfmsg] (M-start |- y) -- ++(1.5, 0) -- ++(0, -0.6) -- node[right, font=\scriptsize, text width=3cm, align=left, xshift=-4cm] {Market orders present?\\\(2701 \geq 2700+1\)? YES} ++(-1.5, 0);
    
    \coordinate (y) at ($(y) + (0, -1.2)$);
    \draw[message, dashed] (M-start |- y) -- node[above, font=\small] {\textbf{UNCROSS}} (S-start |- y);
    
    % Final notes
    \coordinate (y) at ($(y) + (0, -0.8)$);
    \node[note, text width=5cm, fill=red!20, align=center] at ($(T-start |- y)!0.5!(S-start |- y)$) {
        \textbf{Both events occur \break in the same tick}\\
        MIT201 violated 
    };
    
    \end{tikzpicture}
    \caption{Sequence diagram illustrating how the initial conditions found by CodeLogician cause a violation of the MIT201 specification.}
    \label{fig:sequence-diagram}
\end{figure}
\newpage
\section{Multilateral Netting Engine}\label{appendix:mne-gtt}

\subsection{Python model}\
\begin{lstlisting}[language=python]
"""
Multilateral Netting Engine (CCP-style)

This module implements a Central Counterparty (CCP) style multilateral netting engine
that calculates net obligations for participants based on bilateral trades.

Critical Invariants:
1. Zero-Sum (Conservation of Cash): Sum of all net positions = 0.0
2. Netting Efficiency: Sum of |net_positions| <= Sum of |gross_trades|

Regulatory Context: This logic is fundamental to CCPs under EMIR, Dodd-Frank Title VII.
A bug here could cause settlement failures or systemic risk.
"""

from dataclasses import dataclass
from typing import Dict, List


@dataclass
class Trade:
    """
    Represents a bilateral trade between two participants.
    
    Attributes:
        payer_id: The participant who pays (sends cash)
        receiver_id: The participant who receives cash
        amount: The payment amount (must be positive)
    """
    payer_id: str
    receiver_id: str
    amount: float
    
    def __post_init__(self):
        """Validate trade constraints."""
        if self.amount < 0:
            raise ValueError(f"Trade amount must be non-negative, got {self.amount}")
        if self.payer_id == self.receiver_id:
            raise ValueError(f"Self-dealing not allowed: {self.payer_id}")


def calculate_net_obligations(trades: List[Trade]) -> Dict[str, float]:
    """
    Calculate net obligations for all participants in a multilateral netting cycle.
    
    The netting engine aggregates all bilateral trades and computes each participant's
    net position. A positive position means the CCP owes the participant (net receiver);
    a negative position means the participant owes the CCP (net payer).
    
    Args:
        trades: List of bilateral Trade objects
        
    Returns:
        Dictionary mapping participant_id to net_position
        - Positive: Participant is a net receiver (CCP owes them)
        - Negative: Participant is a net payer (they owe the CCP)
        - Zero: Participant is balanced
        
    Invariants Maintained:
        1. Zero-Sum: sum(net_positions.values()) == 0.0
        2. Netting Efficiency: sum(|net|) <= sum(|gross|)
        
    Example:
        >>> trades = [
        ...     Trade("A", "B", 100.0),
        ...     Trade("B", "C", 50.0),
        ...     Trade("C", "A", 30.0)
        ... ]
        >>> net = calculate_net_obligations(trades)
        >>> net
        {'A': -70.0, 'B': 50.0, 'C': 20.0}
        >>> sum(net.values())  # Zero-sum check
        0.0
    """
    net_positions: Dict[str, float] = {}
    
    for trade in trades:
        # Payer: outgoing payment (negative)
        net_positions[trade.payer_id] = net_positions.get(trade.payer_id, 0.0) - trade.amount
        
        # Receiver: incoming payment (positive)
        net_positions[trade.receiver_id] = net_positions.get(trade.receiver_id, 0.0) + trade.amount
    
    return net_positions


def verify_zero_sum(net_positions: Dict[str, float], tolerance: float = 1e-10) -> bool:
    """
    Verify the Zero-Sum invariant (Conservation of Cash).
    
    Args:
        net_positions: Dictionary of participant net positions
        tolerance: Floating-point tolerance for zero comparison
        
    Returns:
        True if sum of all positions is within tolerance of zero
    """
    total = sum(net_positions.values())
    return abs(total) < tolerance


def verify_netting_efficiency(
    trades: List[Trade], 
    net_positions: Dict[str, float]
) -> bool:
    """
    Verify the Netting Efficiency invariant.
    
    The sum of absolute net positions should never exceed the sum of 
    absolute gross trade amounts. Netting should reduce exposure.
    
    Args:
        trades: Original bilateral trades
        net_positions: Computed net positions
        
    Returns:
        True if netting reduces or maintains total exposure
    """
    gross_exposure = sum(trade.amount for trade in trades)
    net_exposure = sum(abs(position) for position in net_positions.values())
    
    return net_exposure <= gross_exposure


def calculate_netting_benefit(
    trades: List[Trade], 
    net_positions: Dict[str, float]
) -> float:
    """
    Calculate the reduction in cash flows achieved by netting.
    
    Returns:
        Percentage reduction in gross obligations (0.0 to 100.0)
    """
    if not trades:
        return 0.0
    
    gross_exposure = sum(trade.amount for trade in trades)
    net_exposure = sum(abs(position) for position in net_positions.values())
    
    if gross_exposure == 0.0:
        return 0.0
    
    reduction = ((gross_exposure - net_exposure) / gross_exposure) * 100.0
    return reduction


\end{lstlisting}

% Appendix: Detailed Case Study - Transportation Risk Manager
% This section provides a concrete example illustrating the aggregate findings from the main evaluation.

\subsection{Evaluation Example: Transportation Risk Manager}
\label{appendix:transportation-case-study}

To illustrate the patterns observed across the 50-model evaluation, we present a detailed analysis of the \texttt{transportation\_risk\_manager} model. This system evaluates route decisions based on transport mode (Truck, Rail, Air, Sea), cargo type (Standard, Fragile, Hazardous, Perishable), weather conditions, and geopolitical risks, determining whether routes are approved and whether emergency routing should be activated.

\subsubsection{The Three Questions}

The model was evaluated using three questions designed to test different aspects of code reasoning:

\begin{enumerate}
  \item \textbf{State Space Estimation (Q1):} ``How many distinct routing scenarios exist for Sea transport when operating under geopolitical tensions (Tensions, Sanctions, or WarZone)?''

  \item \textbf{Conditional Analysis (Q2):} ``Can Air transport ever be rejected for Standard (non-Hazardous) cargo? If yes, under what conditions?''

  \item \textbf{Property Verification (Q3):} ``For Air transport specifically, does having both \texttt{cargo\_value > 100000} AND \texttt{time\_critical < 24} always activate emergency routing when the route is approved?''
\end{enumerate}

\subsubsection{Question 1: State Space Estimation}

\paragraph{CodeLogician Result.}
Region decomposition identified \textbf{117 distinct scenarios} for Sea transport under geopolitical tensions. These scenarios arise from the interaction of:
\begin{itemize}
  \item 3 geopolitical risk levels (Tensions, Sanctions, WarZone)
  \item 5 weather conditions (Clear, Rain, Snow, Storm, Hurricane)
  \item 4 cargo types (Standard, Fragile, Hazardous, Perishable)
  \item Numeric threshold boundaries for \texttt{cargo\_value} and \texttt{time\_critical}
\end{itemize}

Critically, the decomposition reveals that numeric parameters create additional scenario boundaries. For example, one region is defined by:
\begin{lstlisting}[basicstyle=\small\ttfamily]
mode = Sea
geopolitical = Sanctions
cargo = Hazardous
weather = Clear
time_critical <= 23
cargo_value >= 100001
\end{lstlisting}
This scenario results in \texttt{emergency\_routing = true} because three of the four emergency conditions are satisfied. A different region with \texttt{cargo\_value <= 100000} produces \texttt{emergency\_routing = false}---a behaviorally distinct scenario that categorical analysis misses.

\paragraph{LLM Estimates.}
Table~\ref{tab:q1-estimates} summarizes the LLM estimates. All models significantly underestimated the scenario count, with errors ranging from 2$\times$ to 39$\times$.

\begin{table}[h]
\centering
\begin{tabular}{lrrr}
\toprule
\textbf{Model} & \textbf{Estimate} & \textbf{Actual} & \textbf{Error Factor} \\
\midrule
grok-code-fast-1     & 15  & 117 & 7.8$\times$ \\
claude-sonnet-4.5    & 14  & 117 & 8.4$\times$ \\
claude-opus-4.5      & 15  & 117 & 7.8$\times$ \\
gemini-3-pro-preview & 60  & 117 & 2.0$\times$ \\
gpt-5.2              & 3   & 117 & 39$\times$ \\
\bottomrule
\end{tabular}
\caption{Q1 state space estimates by model. All LLMs underestimated the scenario count, consistent with the 0.186 mean State Space Estimation Accuracy reported in the main evaluation.}
\label{tab:q1-estimates}
\end{table}

\paragraph{Analysis.}
The LLMs applied simple categorical reasoning: $3 \text{ (geopolitical)} \times 5 \text{ (weather)} = 15$ scenarios, or variations thereof. This approach correctly counts the categorical combinations but fails to account for how numeric thresholds (\texttt{cargo\_value > 100000}, \texttt{time\_critical < 24}, \texttt{route\_cost < 1000}, \texttt{route\_time < 72}) partition the state space further. The emergency routing logic alone creates multiple behavioral regions within each categorical combination.

GPT-5.2's estimate of 3 scenarios (counting only the geopolitical risk levels) represents an extreme case of categorical oversimplification. Gemini-3-pro's estimate of 60 came closest by attempting to include cargo types ($3 \times 5 \times 4 = 60$), but still missed the numeric threshold boundaries.

\subsubsection{Question 2: Conditional Analysis}

\paragraph{CodeLogician Result.}
Decomposition confirmed that Air transport \textbf{can} be rejected for Standard cargo under exactly \textbf{2 scenarios}:
\begin{enumerate}
  \item \texttt{weather = Hurricane} AND \texttt{geopolitical = WarZone} (extreme risk)
  \item \texttt{weather = Hurricane} AND \texttt{geopolitical = Sanctions} (air grounded)
\end{enumerate}

\paragraph{LLM Results.}
All five models correctly identified both rejection scenarios. For example, Claude Opus 4.5 reasoned:

\begin{quote}
\small
``Air transport can still be rejected via the \texttt{\_is\_route\_viable} check. The \texttt{air\_grounded} condition triggers when: \texttt{mode=AIR AND weather=HURRICANE AND geopolitical=SANCTIONS}. The \texttt{extreme\_risk} condition triggers when: \texttt{weather=HURRICANE AND geopolitical=WAR\_ZONE} (applies to all modes including Air).''
\end{quote}

\paragraph{Analysis.}
This question tests explicit boolean logic that is directly visible in the code:
\begin{lstlisting}[language=Python,basicstyle=\small\ttfamily]
def _is_route_viable(mode, weather, geopolitical):
    extreme_risk = (weather == HURRICANE
                    and geopolitical == WAR_ZONE)
    air_grounded = (mode == AIR
                    and weather == HURRICANE
                    and geopolitical == SANCTIONS)
    return not (extreme_risk or air_grounded)
\end{lstlisting}

The conditions are explicit, categorical, and do not involve numeric thresholds or complex interactions. This explains the 100\% success rate: LLMs excel at tracing explicit control flow when conditions are directly stated in the code. This result is consistent with the relatively higher Control Flow Understanding score (0.746) reported in the main evaluation.

\subsubsection{Question 3: Property Verification}

\paragraph{The Property.}
The question asks whether, for approved Air routes, the conjunction of \texttt{cargo\_value > 100000} AND \texttt{time\_critical < 24} is sufficient to guarantee emergency routing activation.

\paragraph{CodeLogician Result.}
ImandraX \textbf{proved} the property always holds. The intuition behind why this property holds can be understood as follows:

\begin{tcolorbox}[colback=gray!5,colframe=gray!50,title=Why the Property Holds]
Emergency routing activates when at least 3 of 4 conditions are true:
\begin{itemize}
  \item \texttt{c1}: \texttt{cargo\_value > 100000} --- given as true by the property premise
  \item \texttt{c2}: \texttt{time\_critical < 24} --- given as true by the property premise
  \item \texttt{c3}: \texttt{route\_cost < 1000} --- varies depending on weather and geopolitical factors
  \item \texttt{c4}: \texttt{route\_time < 72} --- for Air: $\texttt{base\_time} = 8$, $\texttt{max\_multiplier} = 3$, so $\texttt{max\_time} = 24 < 72$
\end{itemize}

Since \texttt{c4} is always true for Air transport (route time cannot exceed 24 hours), and \texttt{c1} and \texttt{c2} are given by the premise, we have at least 3 conditions satisfied regardless of \texttt{c3}.
\end{tcolorbox}

\paragraph{LLM Results.}
The models produced mixed results:

\begin{itemize}
  \item \textbf{Correct:} GPT-5.2, Gemini-3-pro-preview correctly concluded the property holds.
  \item \textbf{Incorrect:} Grok-code-fast-1 concluded ``No,'' claiming that \texttt{route\_cost = 1000} (exactly) would make \texttt{c3} false, leaving only 2 conditions met.
  \item \textbf{Self-corrected:} Claude Opus 4.5 initially answered ``No'' but corrected itself mid-reasoning.
\end{itemize}

\paragraph{Anatomy of Self-Correction.}
Claude Opus 4.5's reasoning demonstrates the challenge of discovering numeric invariants without formal analysis:

\begin{quote}
\small
``Air route\_cost can exceed 1000: e.g., AIR + STORM + WAR\_ZONE = 500 $\times$ 3 $\times$ 4 = 6000 > 1000. In scenarios where cost $\geq$ 1000 (c3 false) and route\_time $\geq$ 72 (never happens for Air, max is 24), emergency would fail... Wait, re-checking: Air + HURRICANE scenarios are limited. Let me verify c4 is always true: max Air time = 8 $\times$ 3 = 24 < 72, so yes c4 is always true for Air. Correction: With cargo\_value > 100000 (c1=true), time\_critical < 24 (c2=true), and route\_time always < 72 for Air (c4=true), we have 3 conditions, so emergency IS always activated.''
\end{quote}

The model eventually discovered the invariant through explicit arithmetic verification, but only after initially reaching the wrong conclusion. Grok-code-fast-1 made a similar initial error but did not self-correct, focusing on the boundary case \texttt{route\_cost = 1000} without recognizing that \texttt{c4} compensates.

\paragraph{Analysis.}
This question tests the ability to reason about numeric constraints that determine behavior. The key insight (Air's route time is bounded at 24 hours) requires:
\begin{enumerate}
  \item Recognizing that \texttt{route\_time} depends on \texttt{base\_time} and weather multipliers
  \item Computing the maximum possible value for Air specifically
  \item Connecting this bound to the emergency threshold (72 hours)
\end{enumerate}

ImandraX establishes this property through logical deduction. LLMs may arrive at the same conclusion through explicit arithmetic (as Claude Opus eventually did), but this process is unreliable---some models self-correct while others do not, as demonstrated by Grok's failure to recognize that \texttt{c4} compensates for any value of \texttt{c3}.

\subsubsection{Summary}

This case study illustrates the patterns observed across the full evaluation:

\begin{enumerate}
  \item \textbf{State Space Estimation}: LLMs systematically underestimate scenario counts by treating numeric parameters as irrelevant. The 2--39$\times$ error range in Q1 is consistent with the 0.186 mean accuracy reported in the main section.

  \item \textbf{Explicit Control Flow}: When conditions are directly visible as boolean expressions (Q2), LLMs perform well, achieving 100\% accuracy on the conditional analysis task.

  \item \textbf{Numeric Invariants}: Properties requiring discovery of domain-specific bounds (Q3) challenge LLMs. While self-correction is possible, it is unreliable---some models correct themselves, others do not.
\end{enumerate}

While different LLMs exhibit varying performance levels---with some models like Claude Opus 4.5 demonstrating self-correction capabilities that others lack---all models share inherent limitations when reasoning about code without formal methods support. Region decomposition provides exhaustive state space analysis that no LLM achieved, and theorem proving establishes properties with logical certainty that LLMs can only approximate through heuristic reasoning. These limitations are fundamental to current LLM architectures rather than artifacts of any particular model.

\bibliographystyle{abbrvnat}
\renewcommand{\bibsection}{\section*{References}}
 \bibliography{main}
\end{document}